%% file: ICE_revision_V02.tex
\documentclass{article} % For LaTeX2e
\usepackage{iclr2020_conference,times}

\input{math_commands.tex}

\usepackage{hyperref}
\usepackage{url}

\usepackage{dashrule}

\usepackage{algorithm}
\usepackage{algorithmic}

\usepackage{graphicx}
\usepackage{subcaption}
\usepackage[T1]{fontenc}
\usepackage{makecell}

\graphicspath{{Figures/}}
\usepackage{wrapfig}

\usepackage{amsmath,amssymb,amsfonts} 
\usepackage{fixltx2e}
\usepackage{multirow}
\usepackage{booktabs}

\usepackage{etoolbox}
\makeatletter
\patchcmd\@combinedblfloats{\box\@outputbox}{\unvbox\@outputbox}{}{%
	\errmessage{\noexpand\@combinedblfloats could not be patched}%
}%
\makeatother
\title{Instance Cross Entropy for Deep Metric Learning}

\author{%
	Xinshao Wang
	\\
	Anyvision Research Team\\ Queen's University Belfast\\
	\texttt{xwang39@qub.ac.uk} \\
	\AND
	Elyor Kodirov \\
	Anyvision Research Team \\
	\texttt{elyor@anyvision.co} \\
	\And
	Yang Hua \\
	Queen's University Belfast\\
	\texttt{y.hua@qub.ac.uk} \\
	\And
	Neil M. Robertson \\
	Anyvision Research Team\\ Queen's University Belfast \\
	\texttt{n.robertson@qub.ac.uk} \\
}

\iclrfinalcopy % Uncomment for camera-ready version, but NOT for submission.
\begin{document}
	
	\maketitle

	\begin{abstract}
		\vspace{-0.06cm}
		%\textit{Task: }
		Loss functions play a crucial role in deep metric learning thus a variety of them have been proposed. Some supervise the learning process by pairwise or triplet-wise similarity constraints while others take advantage of structured similarity information among multiple data points. 
		In this work,
		we approach deep metric learning from a novel perspective. 
		We propose instance cross entropy (ICE) which 
		measures the difference %means the cross entropy
		between an estimated instance-level matching distribution and its ground-truth one. 
		%
		%We take in the key idea of categorical cross entropy (CCE) due to its natural probabilistic interpretation. 
		%
		ICE has three main appealing properties. 
		Firstly, similar to categorical cross entropy (CCE), ICE has clear probabilistic interpretation and exploits structured semantic similarity information for learning supervision. 
		Secondly, ICE is scalable to infinite training data as it learns on mini-batches iteratively and is independent of the training set size. 
		%complying with the intrinsic requirement of metric learning. 
		Thirdly, motivated by our relative weight analysis, seamless sample reweighting is incorporated. It rescales samples' gradients to control the differentiation degree over training examples instead of truncating them by sample mining.  
		In addition to its simplicity and intuitiveness, extensive experiments on three real-world benchmarks demonstrate the superiority of ICE. 
		%, e.g., outperforming the state-of-the-art by 2.1\% in terms of Recall@1 on SOP \citep{song2016deep}.     
	\end{abstract}
	
	\vspace{-0.16cm}
	\section{Introduction}
	\label{introduction}
	\vspace{-0.16cm}
	
	Deep metric learning (DML) aims to learn a non-linear embedding function (a.k.a. distance metric) such that the semantic similarities over samples are well captured in the feature space \citep{tadmor2016learning,sohn2016improved}.
	%Deep metric learning (DML) aims to learn a deep distance metric which is consistent with the semantic similarity of samples \citep{schroff2015facenet,song2016deep}.    
	Due to its fundamental function of learning discriminative representations, DML has diverse applications, such as image retrieval \citep{song2016deep}, clustering \citep{song2017deep}, verification \citep{schroff2015facenet}, few-shot learning \citep{vinyals2016matching} and zero-shot learning \citep{bucher2016improving}. 
	
	\vspace{-0.06cm}
	A key to DML is to design an effective and efficient loss function for supervising the learning process, thus significant efforts have been made \citep{chopra2005learning,schroff2015facenet,sohn2016improved,song2016deep,song2017deep,law2017deep,wu2017sampling}. 
	%Specifically, some loss functions follow stringent data preparation protocol, and they learn the metric from pairs or triplets or n-pairs relational constraints \citep{chopra2005learning,hadsell2006dimensionality,schroff2015facenet,bell2015learning}. 
	Some loss functions learn the embedding function from pairwise or triplet-wise relationship constraints \citep{chopra2005learning,schroff2015facenet,tadmor2016learning}.
	However, they are known to not only suffer from an increasing number of non-informative samples during training, but also incur considering only several instances per loss computation. Therefore, informative sample mining strategies are proposed \citep{schroff2015facenet,wu2017sampling,wang2019deep}.  
	Recently, several methods consider semantic relations among multiple examples 
	%within a mini-batch 
	to exploit their similarity structure \citep{sohn2016improved,song2016deep,song2017deep,law2017deep}. Consequently, these structured losses achieve better performance than pairwise and triple-wise approaches.

	% On the one hand, some algorithms learn from pairwise or triplet-wise relation constraints \citep{chopra2005learning,hadsell2006dimensionality,schroff2015facenet,bell2015learning}.  The loss functions based on pairwise or triplet-wise relations are known to suffer from an increasing number of trivial (non-informative) samples as the model improves during training. Therefore, effective and efficient sample mining strategies are popular \citep{schroff2015facenet,wu2017sampling, wang2019deep}.  On the other hand, some methods consider relations among multiple data pairs and make use of the global similarity structure of the input data \citep{song2017deep,law2017deep}. They achieve much better performance than pairwise or triple-wise approaches.   
	\vspace{-0.06cm}
	In this paper, we tackle the DML problem from a novel perspective. Specifically, we propose a novel loss function inspired by CCE. CCE is well-known in classification problems owing to the fact that it has an intuitive probabilistic interpretation and achieves great performance, e.g., ImageNet classification \citep{russakovsky2015imagenet}. 
	However, since CCE learns a decision function which predicts the class label of an input, it learns class-level centres for reference \citep{zhang2018heated,wang2017normface}. Therefore, CCE is not scalable to infinite classes and cannot generalise well when it is directly applied to DML \citep{law2017deep}.    
	%However, since the objective of metric learning and classification are different i.e., learning a decision function vs. metric CCE cannot be applied to learning a deep metric directly. Moreover, CCA fails if the number of classes are infinite, whereas DML methods succeeds. 
	
\begin{figure}[!t]
		%\centering
		\begin{subfigure}[h]{0.46\textwidth}
			%\centering
			\vspace{-0.16cm}
			\includegraphics[width=1.01\textwidth]{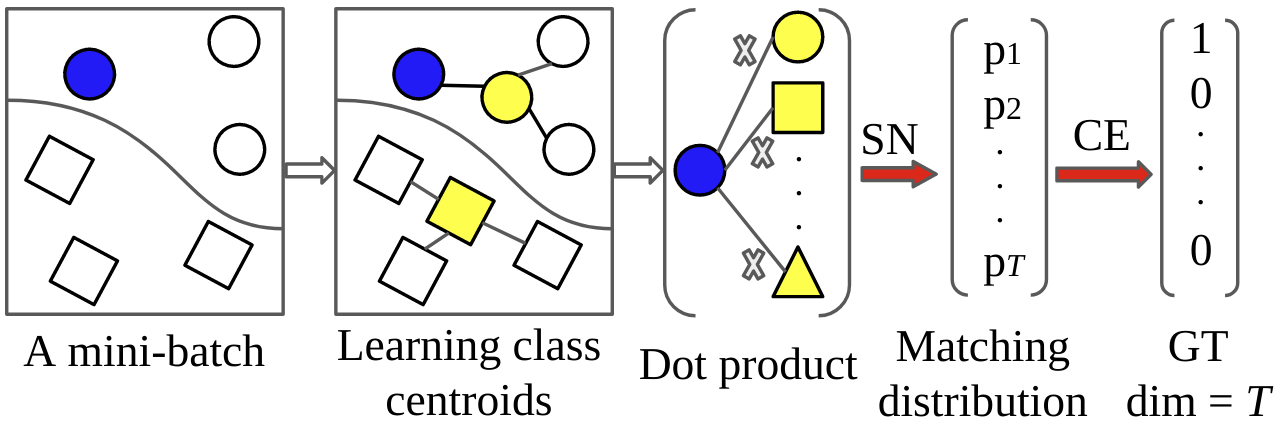}
			\captionsetup{width=0.98\textwidth}
			\caption{
				A query versus {\textbf{learned parametric class centroids}}. 
				{All $T$ classes in the training set are considered.}
				Prior work: CCE, Heated-up \citep{zhang2018heated}, NormFace \citep{wang2017normface}. 
			}
			\label{fig:Learned_Center}
			%\vspace{0.05cm}
		\end{subfigure}
		%\hspace*{-3cm}
		~~~~~~
		\begin{subfigure}[h]{0.5\textwidth}
			\vspace{-0.16cm}
			\includegraphics[width=0.96\textwidth]{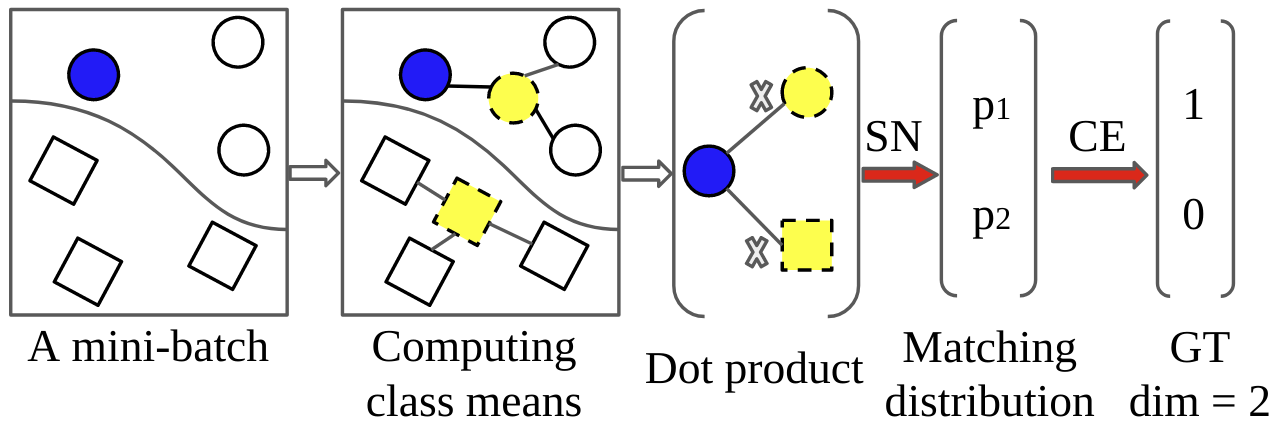}
			\captionsetup{width=1.0\textwidth}
			\caption{
				A query versus \textbf{{non-parametric class means}}. 
				{Only classes in the mini-batch are considered.}
				Representative work: {TADAM} \citep{oreshkin2018tadam},  {DRPR} \citep{law2019dimensionality}, {Prototypical Networks} \citep{snell2017prototypical}.
				}
			\label{fig:Mean_Center}
			%\vspace{0.05cm}
		\end{subfigure}
		
		\vspace{0.06cm}
		\hdashrule{1.0\textwidth}{1pt}{1.0pt}
		\vspace{0.06cm}
		\begin{subfigure}[h]{0.46\textwidth}
			%\vspace{-0.06cm}
			\includegraphics[width=0.85\textwidth]{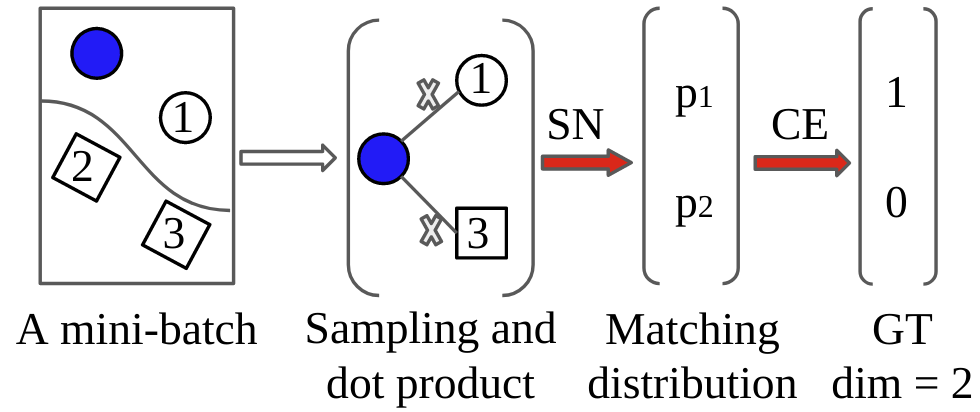}
			\captionsetup{width=1.0\linewidth}
			\caption{\textit{N}-pair-mc \citep{sohn2016improved}: A query versus {\textbf{one instance per class}}. A mini-batch has to be 2 examples per class rigidly. Only one instance per negative class is randomly sampled out of 2. 
			}
			\label{fig:NPairMC}
		\end{subfigure}
		~~~~~~
		\begin{subfigure}[h]{0.5\textwidth}
			%\vspace{-0.10cm}
			\includegraphics[width=0.98\textwidth]{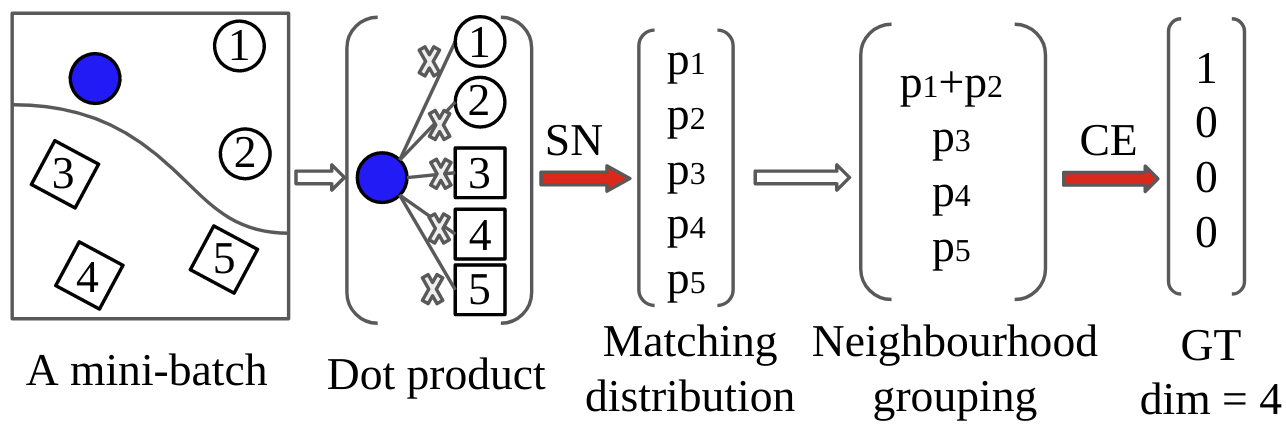}
			\captionsetup{width=1.0\linewidth}
			\caption{
				NCA \citep{goldberger2005neighbourhood} and S-NCA \citep{wu2018improving}: A query versus \textbf{the rest instances}. $\hphantom{dddddddddddddddddddddddddddddddddddddddddddddddddddddddddddddddddddddddddddddddddddddddddddddddddddddddddddddddddddddddddddddddddddddddddddddddddddddddddddddddddddddddddddd}$
			}
			\label{fig:NCA}
		\end{subfigure}
		
		\vspace{0.06cm}
		\hdashrule{1.0\textwidth}{1pt}{1.0pt}
		\begin{subfigure}[h]{0.46\textwidth}
			\vspace{-0.20cm}
			\centering
			\includegraphics[width=0.86\textwidth]{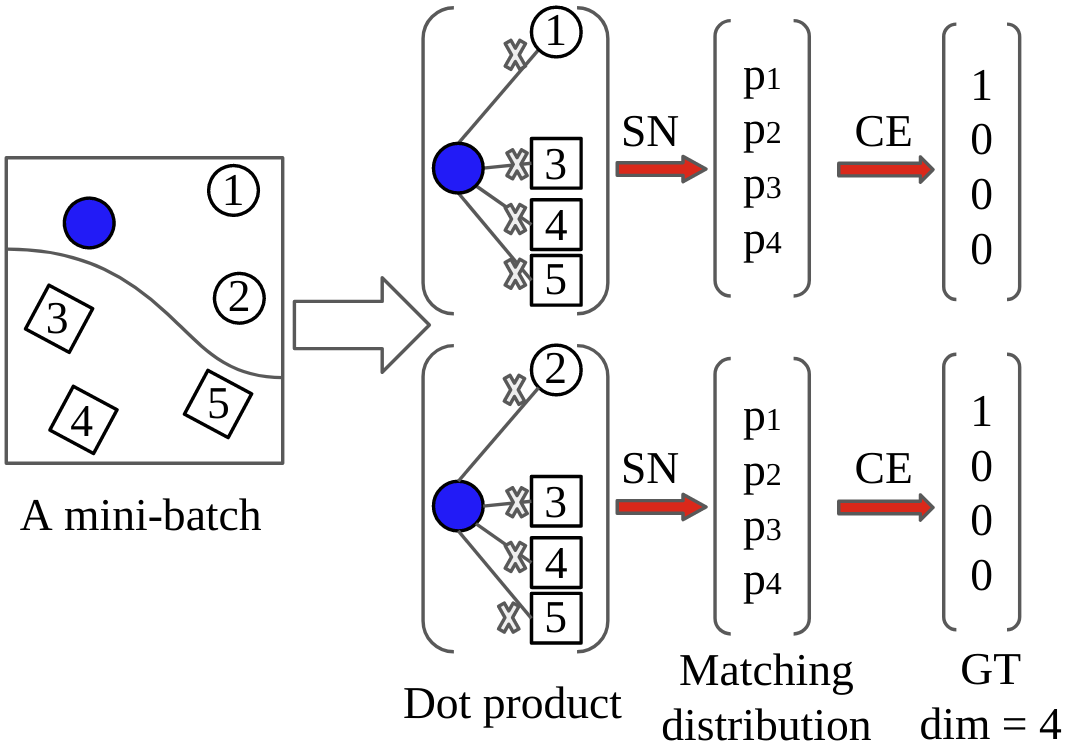}
			\captionsetup{width=1.0\textwidth}
			\caption{
				Our ICE: A query versus \textbf{one positive and all negatives per distribution}. A query's number of matching distributions is defined by the number of its positive examples. 
				%The dimension of GT is `3 (number of negative instances)' + `1 (one positive instance)'.  
			}
			\label{fig:ICE}
			%\vspace{0.05cm}
		\end{subfigure}
		~~~~~~
		\begin{subfigure}[h]{0.5\textwidth}
			~~~~~~~~
			\vspace{-0.08cm}
			\includegraphics[width=0.86\textwidth]{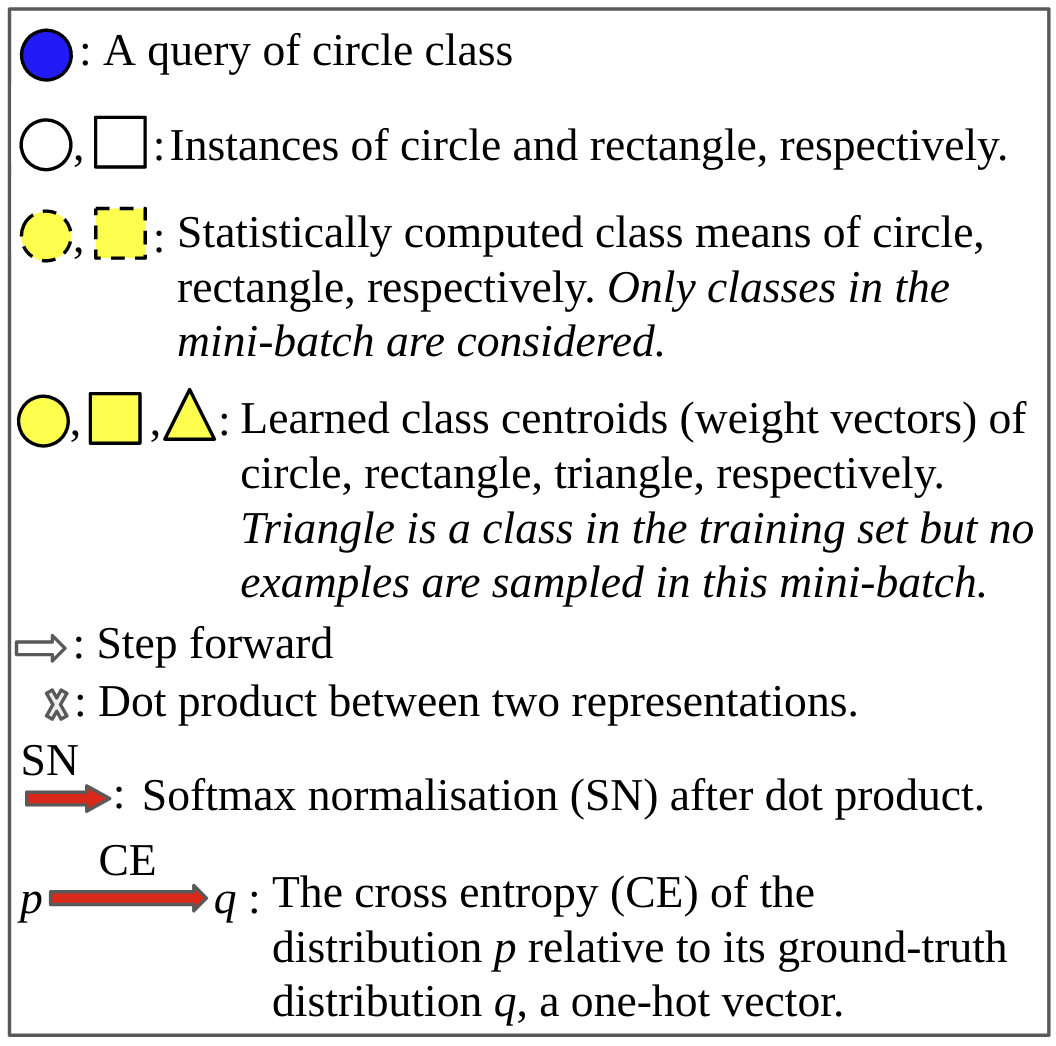}
			%\captionsetup{width=1.1\linewidth}
		\end{subfigure}
	
		\caption{
			Our ICE and related losses. The first row shows prior work of {a query versus class centres/means} while the second row displays the work of {a query versus instances.}  
			Note that the cross entropy computation and interpretation are different in different losses.	
			For a mini-batch, we show two classes, i.e., circle and rectangle, with 3 examples per class except \textit{N}-pair-mc which requires 2 samples per class. 
			%The blue circle is a query (anchor).  
			The icons are at the right bottom. GT means ground-truth matching distribution. 
			When illustrating the losses of a query versus instances in (c), (d) and (e), we index those instances with numbers for clarity, except the query.
			%Best viewed in colour. 
		}
		\label{fig:comparing_losses}
		\vspace{-0.3cm}
	\end{figure}

	\vspace{-0.06cm}
	With scalability and structured information in mind, we introduce instance cross entropy (ICE) for DML. It learns an embedding function by minimising the cross entropy between a predicted instance-level matching distribution and its corresponding ground-truth. 
	In comparison with CCE, given a query, CCE aims to maximise its \textit{matching probability with the class-level context vector} (weight vector) of its ground-truth class, whereas ICE targets at maximising its \textit{matching probability with it similar instances}. As ICE does not learn class-level context vectors, it is scalable to infinite training classes, which is an intrinsic demand of DML. 
	Similar to \citep{sohn2016improved,song2016deep,song2017deep,law2017deep,goldberger2005neighbourhood,wu2018improving}, ICE is a structured loss as it also considers all other instances in the mini-batch of a given query. 
	We illustrate ICE with comparison to other structured losses in Figure~\ref{fig:comparing_losses}.
	
	\vspace{-0.06cm}
	A common challenge of instance-based losses is that many training examples become trivial as model improves.
	Therefore, 
	%beyond scalability and structured supervision, 
	we integrate seamless sample reweighting into ICE, which functions similarly with various sample mining schemes \citep{sohn2016improved,schroff2015facenet,shi2016embedding,yuan2017hard,wu2017sampling}. Existing mining methods require either separate time-consuming process, e.g., class mining \citep{sohn2016improved}, or distance thresholds for data pruning \citep{schroff2015facenet,shi2016embedding,yuan2017hard,wu2017sampling}. Instead, our reweighting scheme works without explicit data truncation and mining. It is motivated by the relative weight analysis between two examples. The current common practice of DML is to learn an angular embedding space by projecting all features to a unit hypersphere surface \citep{song2017deep,law2017deep,movshovitz2017no}.  We identify the challenge that without sample mining, informative training examples cannot be differentiated and emphasised properly because the relative weight between two samples is strictly bounded. We address it by sample reweighting, which rescales samples' gradient to control the differentiation degree among them. 
	
	\vspace{-0.06cm}
	Finally, for intraclass compactness and interclass separability, most methods \citep{schroff2015facenet,song2016deep,tadmor2016learning,wu2017sampling} use distance thresholds to decrease intraclass variances and increase interclass distances. In contrast, we achieve the target from \textit{a perspective of instance-level matching probability}. \textit{Without any distance margin constraint}, ICE makes no assumptions about the boundaries between different classes. Therefore, ICE is easier to apply in applications where we have no prior knowledge about intraclass variances.    
	
	\vspace{-0.06cm}
	Our contributions are summarised: (1) We approach DML from a novel perspective by taking in the key idea of matching probability in CCE. We introduce ICE, which is scalable to an infinite number of training classes and exploits structured information for learning supervision. 
	(2) A seamless sample reweighting scheme is derived for ICE to address the challenge of learning an embedding subspace by projecting all features to a unit hypersphere surface.   
	(3) We show the superiority of ICE by comparing with state-of-the-art methods on three real-world datasets.

	\vspace{-0.06cm}
	%\vspace{-0.1cm}
	\section{Related Work}
	\vspace{-0.16cm}
	%%%%%%%%%%%%%%%%%%%%%%%%%%%%%%%%%%%%%%%%%%%%%%
	%In section \ref{sec:minimising_ICE}, we have analysed the relationships of our method to CCE and sample mining strategies. 
	%Here we systematically analyse the connections of  our ICE to other losses and sample mining. 
	%	We compare ICE with related structured loss functions in Figure~\ref{fig:comparing_losses}. We group them into two: instance-based and proxy-based methods. %More details are as follows.

	%%%%%%%%%%%%%%%%%%%%%%%%%%%%%%%%%%%%%%%%%%%%%%%%%%%%%
	
	\vspace{-0.05cm}
	\subsection{Structured Losses by Query versus Class Centres}
	\vspace{-0.16cm}
	\textbf{Heated-up}, \textbf{NormFace}, \textbf{TADAM},  \textbf{DRPR}, \textbf{Prototypical Networks}, \textbf{Proxy-NCA}. 
	These methods calculate the {similarities between a query and class centres (a.k.a. proxies or prototypes) instead of other instances} \citep{zhang2018heated,wang2017normface,oreshkin2018tadam,law2019dimensionality,snell2017prototypical,movshovitz2017no}. 
	In Heated-up and NormFace, the class centres are learned parameters of a fully connected layer, which is similar to Center Loss \citep{wen2016discriminative}. While in TADAM, DRPR, and Prototypical Networks, a class centre is the mean over all embeddings of a class.  
	By comparing a sample with other examples other than class centres, more informative 
	instances
	%ones in a class 
	can contribute more in ICE.  
	%%%%%%%%%%%%%%%%%%%%%%%%%%%%%%%%%%%%%%%%%%%%%%%%%%%%%
	%\emph{\textbf{Proxy-NCA}} \citep{movshovitz2017no} aims to address the sampling problem using proxies.
	%It is the traditional NCA loss defined over proxies, which approximates the original data points:
	%\begin{equation}
	%L(\mathbf{a},\mathbf{u},\mathbf{Z})=-\log(
	%\frac{\exp(-d(\mathbf{a},p(\mathbf{u}))))}
	%{\sum_{\mathbf{z} \in \mathbf{Z}} \exp(-d(\mathbf{a},p(\mathbf{z})))}
	%).
	%\end{equation}
	%Here $\mathbf{a}$ is an anchor, $\mathbf{Z}$ is its negative set, $p(\mathbf{u})$ and $p(\mathbf{z})$ are the proxies of its positive and negative points, respectively.  and $d(\cdot, \cdot)$ computes the Euclidean distance between two examples. 
	%%
	%\emph{Proxy-NCA does not fulfil the scalability requirement of DML} as it needs to learn the proxy for every class.

	%%%%%%%%%%%%%%%%%%%%%%%%%%%%%%%%%%%%%%%%%%%%%%%%%%
	%\vspace{-0.01cm}
	\vspace{-0.14cm}
	\subsection{Structured Losses by Query versus Instances}
	\vspace{-0.16cm}	
	%We compare ICE with related structured loss functions in Figure~\ref{fig:comparing_losses}. More details are as follows. 
	
	%NCA aims to optimise the probability of all similar instances being correctly recognised:
	%\begin{equation}
	%\label{equation:NCA}
	%\begin{aligned}
	%L(\{(\mathbf{x}_i, y_i)\}_{i=1}^{N};f)
	%= &  
	%\frac{1}{2N} 
	%\sum_{a=1}^{N} -\log
	%\frac{
	%	\sum_{j \neq a:y_j=y_a} \exp(\mathbf{f}_a^\top \mathbf{f}_j)
	%}
	%{\sum_{k \neq a} \exp(\mathbf{f}_a^\top \mathbf{f}_k)}.
	%\end{aligned}
	%\end{equation} 
	
	\textbf{NCA} \citep{goldberger2005neighbourhood}, \textbf{S-NCA} \citep{wu2018improving}. NCA learns similarity relationships between instances. 
	Since original NCA learns the whole training data and its time complexity is quadratically proportional to the scale of training data, S-NCA is proposed recently with linear time complexity with respect to the training data size. Instead, ICE is scalable to infinite training data by iterative learning on randomly sampled small-scale instances matching tasks.
	S-NCA and NCA share the same learning objective. However, they treat the event of all similar instance being correctly recognised \textit{as a whole} by a sum accumulator. Instead, we maximise the probability of every similar sample being correctly identified \textit{individually}. 
	Therefore, ICE's optimisation task is harder, leading to better generalisation.% consequently.
	
	%%%%%%%%%%%%%%%%%%%%%%%%%%%%%%%%%%%%%%%%%%%%%%%%%%%%%
	\vspace{-0.08cm}
	\textbf{\textit{N}-pair-mc} \citep{sohn2016improved}. The aim of \textit{N}-pair-mc is to {identify one positive example from $N-1$ negative examples of $N-1$ classes} (one negative example per class).
	%:
	%%%%\vspace{-10pt}
	%\begin{equation}
	%\begin{aligned}
	%L(\{(\mathbf{x}_i^1,\mathbf{x}_i^2)\}_{i=1}^N;f) = &  
	%\frac{1}{N} 
	%\sum_{i=1}^{N} -\log
	%\frac{\mathbf{f}_i^{1\top} \mathbf{f}_i^2}
	%{\sum_{j} \exp(\mathbf{f}_i^{1\top} \mathbf{f}_j^{2})}
	%\end{aligned},
	%\end{equation}
	%
	%where $ \{(\mathbf{x}_i^1, \mathbf{x}_i^2)\}_{i=1}^N$ are $N$ pairs of examples from $N$ different classes, i.e., $y_i \neq y_j, \forall i \neq j$. Here, $\mathbf{x}_i^1$ and $\mathbf{x}_i^2$ are two images from the same class.
	%
	In other words, only one positive and one negative instance per class are considered per loss computation by simulating CCE exactly. Instead, ICE exploits all negative examples to benefit from richer information. 
	When constructing mini-batches, \textit{N}-pair-mc requires expensive offline class mining and samples 2 images per class.   
	%%%%%%%%%%%%%%%%%%%%%%%%%%%%%%%%%%%%%%%%%%%%%%%%%%%%%
	According to \citep{sohn2016improved} \textit{N}-pair-mc is superior to NCA. 
	%, while ICE outperforms \textit{N}-pair-mc.
	%In \textit{N}-pair-mc \citep{sohn2016improved}, NCA is implemented for deep metric learning but its performance is not as good as \textit{N}-pair-mc.
	%%%%%%%%%%%%%%%%%%%%%%%%%%%%%%%%%%%%%%%%%%%%%%

	%%%%%%%%%%%%%%%%%%%%%%%%%%%%%%%%%%%%%%%%%%%%%%
	\vspace{-0.06cm}
	\textbf{Hyperbolic} \citep{nickel2018learning}. It aims to preserve the similarity structures among instances as well. However, it learns a hyperbolic embedding space where the distance depends only on norm of embeddings. Instead, we learn an angular space where the similarity depends only on the angle between embeddings. Besides, Hyperbolic requires a separate sampling of semantic subtrees when the dataset is large. 
	
	%%%%%%%%%%%%%%%%%%%%%%%%%%%%%%%%%%%%%%%%%%%%%%

	%%%%%%%%%%%%%%%%%%%%%%%%%%%%%%%%%%%%%%%%%%%%%%%%%%%%%
	\vspace{-0.16cm}
	\subsection{Sample Mining and Weighting}
	\label{sec:connect_to_OSM}
	\vspace{-0.16cm}
	
	Mining informative examples or emphasising on them are popular strategies in DML: 1) 
	Mining non-trivial samples during training is crucial for faster convergence and better performance. Therefore, sample mining is widely studied in the literature. 
	In pairwise or triplet-wise approaches \citep{schroff2015facenet,wu2017sampling,huang2016local,yuan2017hard}, data pairs with higher losses are emphasized during gradient backpropagation. 
	As for structured losses, Lifted Struct \citep{song2016deep} also focuses on harder examples. 
	Furthermore, \citep{sohn2016improved} and \citep{suh2019stochastic} propose to mine hard negative classes to construct informative input mini-batches. Proxy-NCA \citep{movshovitz2017no} addresses the sampling problem by learning class proxies. %Proxy-NCA shares similar idea with CCE where class contexts vectors (representatives) are learned. 
	2) Assigning higher weights to informative examples is another effective scheme \citep{wang2019ranked, wang2019multi}.  Beyond, there are some other novel perspectives to address sample mining or weighting, e.g., hardness-aware examples generation \citep{zheng2019hardness} and divide-and-conquer of the embedding space \citep{sanakoyeu2019divide}.
	
	\vspace{-0.06cm}
	Our proposed ICE has a similarity scaling factor which helps to emphasise more on informative examples.  
	Moreover, as described in \citep{schroff2015facenet}, very hard negative pairs are likely to be outliers and it is safer to mine semi-hard ones. 
	In ICE, the  similarity scaling factor  is flexible in that it controls the emphasis degree on harder samples. 
	Therefore, a proper similarity scaling factor can help mine informative examples and alleviate the disturbance of outliers simultaneously. \textit{What makes ours different is that we do not heuristically design the mining or weighting scheme}. Instead, it is built-in and we simply scale it as demonstrated in Section~\ref{sec:implict_weighting_generalisation}. 
	
	\vspace{-0.16cm}
	\subsection{Discussion}
	\vspace{-0.16cm}
	We remark that Prototypical Networks, Matching Networks \citep{vinyals2016matching} and NCA are also scalable and do not require distance thresholds. Therefore, they are illustrated and differentiated in Figure~\ref{fig:comparing_losses}. 
	Matching Networks are designed specifically for one-shot learning. Similarly, \citep{triantafillou2017few} design mAP-SSVM and mAP-DLM for few-shot learning, which directly optimises the retrieval performance mAP when multiple positives exist. 
	FastAP \citep{cakir2019deep} is similar to \citep{triantafillou2017few} and optimises the ranked-based average precision.  
	Instead, ICE processes one positive at a time.
	Beyond, the setting of few-shot learning is different from deep metric learning: Each mini-batch is a complete subtask and contains a support set as training data and a query set as validation data in few-shot learning. Few-shot learning applies episodic training in practice. 
	
	\vspace{-0.06cm}
	Remarkably, TADAM formulates instances versus class centres and also has a metric scaling parameter for adjusting the impact of different class centres. Contrastively, ICE adjusts the influence of other instances. Furthermore, ours is not exactly distance metric scaling since we simply apply naive cosine similarity as the distance metric at the testing stage. %although they look similar in formats.
	That is why we interpret it as a weighting scheme during training. 
	 
	%	There are two types of strategies for mining informative samples. One is hard mining \citep{schroff2015facenet,huang2016local,yuan2017hard,shi2016embedding}, i.e., assigning a binary weight (1/0) to each sample. %, i.e., keeping or dropping it. 
	%	The other one is soft mining \citep{wang2019deep}, i.e., assigning a continuous weight to each sample depending on hardness degree. %  to differentiate samples of different hardness degree.   
	
	%	Our sample reweighting scheme can be regarded as non-linear post-processing of partial derivatives' magnitudes, i.e., a type of online soft mining \citep{wang2019deep}.
	%	
	%	As indicated in Eq.~(\ref{equation:final_weight_pos}) and Eq.~(\ref{equation:final_weight_neg}), the weights of positive and negative points align well with our intuition. On the one hand, for a more difficult positive instance, its rescaled probability $\hat{p}$ is less than 1 and weight is higher. On the other hand, a more difficult negative example has higher matching probability and weight.
	%	% 
	%	By introducing the scaling parameter $s$, we can \textit{control the emphasis degree on harder data points in different applications}. 
	%$s$ is the only hyper-parameter during training and can be easily selected via cross-validation. 
	
	\vspace{-0.16cm}
	\section{Instance Cross Entropy}
	\label{sec:minimising_ICE}
	\vspace{-0.16cm}
	
	{\textbf{Notation}}. $\mathbf{X} = \{(\mathbf{x}_i, y_i)\}_{i=1}^{N} = \{\{\mathbf{x}_i^c\}_{i=1}^{N_c}\}_{c=1}^C$  is an input mini-batch, where $\mathbf{x}_i \in \mathbb{R}^{h\times w\times 3}$ and $y_i \in \{1, ... , C\}$ represent $i$-th image and the corresponding label, respectively; $\{\mathbf{x}_i^c\}_{i=1}^{N_c}$ is a set of $N_c$ images belonging to $c$-th class, $\forall c, N_c \ge 2$. The number of classes $C$ is generally much smaller than the total number of classes $T$ in the training set ($C \ll T$). Note that $T$ is allowed to be extremely large in DML. Given a sufficient number of different mini-batches, our goal is to learn an embedding function $f$ that captures the semantic similarities among samples in the feature space. We represent deep embeddings of X as $\{\{\mathbf{f}_i^c = f(\mathbf{x}_i^c) \}_{i=1}^{N_c}\}_{c=1}^C$.
	Given a query, `positives' and `negatives' refer to samples of the same class and different classes, respectively.
	% in some fine-grained applications, e.g. face recognition \citep{schroff2015facenet}.   
	%$C \ll T$
	%The objective of ICE is to learn a discriminative embedding function $f$ (a.k.a. deep metric) such that any query can be matched correctly with its positive examples.  
	%At least two images exist in every class so that every image can serve as the query iteratively and its ICE loss can be computed. 
	%Given any query, we compute the joint matching probability of it being matched correctly with its all positive samples. Its ICE loss is the negative log-likelihood of the joint matching probability. 
	%Hereafter, we will introduce CCE briefly in \ref{sec:CCE} before presenting our ICE in detail in \ref{sec:ICE}. We explain $L_2$ feature normalisation from the perspective of regularisation in \ref{sec:L2regularisation}. Implicit weighting in the ICE loss is exposed and generalised in \ref{sec:implict_weighting_generalisation}. For any anchor, we normalise the weight of all other instances in the mini-batch, which is named Anchor-based Weight Normalisation and introduced in \ref{sec:anchor_based_weight_normalisation}. Finally, we connect our weight analysis and generalisation to online sample mining in \ref{sec:connect_to_OSM}.  
	
	%	We first revisit categorical cross entropy (CCE) and then proceed to our method, i.e., instance cross entropy (ICE).
	
	\vspace{-0.16cm}
	\subsection{Revisiting Categorical Cross Entropy}
	\label{sec:CCE}
	\vspace{-0.12cm}
	CCE is widely used in a variety of tasks, especially classification problems. As demonstrated in \citep{liu2016large}, a deep classifier consists of two joint components: \textit{deep feature learning} and \textit{linear classifier learning}. 
	%\footnote{$f$ is a predefined embedding network whose parameters need to be learned.}
	The feature learning module is a transformation (i.e., embedding function $f$ ) composed of convolutional and non-linear activation layers. The classifier learning module has one neural layer, which learns $T$ class-level context vectors such that any image has the highest compatibility (logit) with its ground-truth class context vector:
	\vspace{-0.1cm}  
	\begin{equation}
	\label{equation:cce_prob}
	p({\mathbf{w}_{y_i} | \mathbf{x}_i}) = \frac{\exp(\mathbf{f}_i^\top\mathbf{w}_{y_i})}{\sum\nolimits_{k=1}^T \exp(\mathbf{f}_i^\top\mathbf{w}_k)} 
	%= \frac{\exp(\mathbf{f}_i^\top\mathbf{w}_{y_i})}{\exp(\mathbf{f}_i^\top\mathbf{w}_{y_i})+\sum\nolimits_{k \neq y_i} \exp(\mathbf{f}_i^\top\mathbf{w}_k)}
	\text{~~~and~~~}
	L_{\mathrm{CCE}}(\mathbf{X};f, \mathbf{W}) =  -\sum\nolimits_{i=1}^N \log p({\mathbf{w}_{y_i} | \mathbf{x}_i}),
	\end{equation}
	%\vspace{-0.1cm} 
	where $\mathbf{f}_i = f(\mathbf{x}_i) \in \mathbb{R}^d$ is a $d$-dimensional vector, $p({\mathbf{w}_{y_i} | \mathbf{x}_i})$ is the probability (softmax normalised logit) of $\mathbf{x}_i$ matching $\mathbf{w}_{y_i}$, %$\mathbf{w}_k$ is the $k-$th class context vector, which can also be interpreted as an unique class-level representative.
	$\mathbf{W} = \{\mathbf{w}_k \in \mathbb{R}^{d} \}_{k=1}^T$ is the learned parameters of the classifier.
	% (the last layer in a deep CNN network). 
	%
	During training, the goal is to maximise the joint probability of all instances being correctly classified. The identical form is to minimise the negative log-likelihood, i.e., $L_{\mathrm{CCE}}(\mathbf{X};f, \mathbf{W})$.  
	%	\begin{equation}
	%	\begin{aligned}
	%	L_{\mathrm{CCE}}(\mathbf{X};f, \mathbf{W}) &=  -\sum_{i=1}^N \log p({\mathbf{w}_{y_i} | \mathbf{x}_i}) %\\
	%	%&= - \sum_{c=1}^C \sum_{i=1}^{N_c} \log p(c|\mathbf{x}_i^c).
	%	\end{aligned}
	%	\end{equation}
	%	
	Therefore, the learning objective of CCE is:
		\vspace{-0.16cm} 
		%%\vspace{-0.88cm}
		\begin{equation}
		\label{equation:cce_loss}
		%{f, \mathbf{W}} &=
		%\max_{f, \mathbf{W}}   
		\argmax_{f, \mathbf{W}}
		\prod\nolimits_{i=1}^N p({\mathbf{w}_{y_i} | \mathbf{x}_i}) 
		= 
		%\min_{f, \mathbf{W}} 
		\argmin_{f, \mathbf{W}}
		L_{\mathrm{CCE}}(\mathbf{X};f, \mathbf{W}).
		\end{equation}

	\vspace{-0.28cm}
	\subsection{Instance Cross Entropy}
	\label{sec:ICE}
	\vspace{-0.16cm}
	%The overall pipeline of ICE is illustrated in Figure~\ref{fig:pipeline}.
	%CCE is used for classification problem hence the class-context weights are learned. With the proposed ICE, since it is for DML the class-context vectors are unnecessary. 
	In contrast to CCE, ICE is a loss for measuring instance matching quality (lower ICE means higher quality) and does not need class-level context vectors.  
	We remark that an anchor may have multiple positives, which are isolated in separate matching distributions. There is a matching distribution for every anchor-positive pair versus their negatives as displayed in Figure~\ref{fig:ICE}. 
	
	%Specifically, given $\mathbf{X}$ and $f$, we have $\{\{\mathbf{f}_i^c = f(\mathbf{x}_i^c) \}_{i=1}^{N_c}\}_{c=1}^C$. 
	%Note that $C \ll T$ (The total number of classes in the training set, which can be infinite).
	
	\vspace{-0.06cm}
	Let $\mathbf{f}_a^c$ be a random query, we compute its similarities with the remaining points using dot product.
	%as in  \citep{jetley2018learn}. 
	% 
	%Following %, the compatibility between two feature vectors is measured by their dot product. 
	%To this end, 
	We define the probability of the given anchor $\mathbf{x}_a^c$ matching one of its positives  $\mathbf{x}_i^c (i \neq a)$ as follows:
	%Then, for the given anchor $\mathbf{x}_a^c$, the probability of its any positive data point $\mathbf{x}_i^c (i \neq a)$ being identified matching itself is obtained by a softmax normalisation of their compatibility: 
	%\begin{center}
		%%\vspace{-0.8cm}
		\vspace{-0.2cm} 
		\begin{equation}
		\label{equation:ice_prob_pos}
		p(\mathbf{x}_i^c | \mathbf{x}_a^c) = \frac{\exp( {\mathbf{f}_a^c}^\top {\mathbf{f}_i^c} )}
		{	\exp( {\mathbf{f}_a^c}^\top {\mathbf{f}_i^c} ) + 
			\sum\nolimits_{o \neq c}  \sum\nolimits_{j} 
			\exp({ \mathbf{f}_a^c}^\top {\mathbf{f}_j^o} )
		},
		\end{equation}
		\vspace{-0.3cm} 
		%%\vspace{-0.5cm}
	%\end{center}
	%\begin{align}
	%\label{equation:ice_prob_pos}
	%p(\mathbf{x}_i^c | \mathbf{x}_a^c) = \frac{\exp( {\mathbf{f}_a^c}^\top {\mathbf{f}_i^c} )}
	%{	\exp( {\mathbf{f}_a^c}^\top {\mathbf{f}_i^c} ) + 
	%	\sum\limits_{o \neq c}  \sum\limits_{j} 
	%	\exp({ \mathbf{f}_a^c}^\top {\mathbf{f}_j^o} )
	%},
	%\end{align}
	
	where ${\mathbf{f}_a^c}^\top {\mathbf{f}_i^c}$ is the similarity between $\mathbf{x}_a^c$ and $\mathbf{x}_i^c$ in the embedding space, $\sum\nolimits_{o \neq c}  \sum\nolimits_{j} \exp({ \mathbf{f}_a^c}^\top {\mathbf{f}_j^o} )$ is the sum of similarities between $\mathbf{x}_a^c$ and its all negatives. Similarly, when the positive is $\mathbf{x}_i^c$, the probability of one negative point $\mathbf{x}_j^o (o\neq c)$ matching the anchor is:
	%\begin{center}
		%%\vspace{-0.8cm}
		\vspace{-0.2cm} 
		\begin{equation}
		\label{equation:ice_prob_neg}
		p(\mathbf{x}_j^o | \mathbf{x}_a^c,\mathbf{x}_i^c) = \frac{  
			\exp({ \mathbf{f}_a^c}^\top {\mathbf{f}_j^o})  }
		{	\exp( {\mathbf{f}_a^c}^\top {\mathbf{f}_i^c} ) + 
			\sum\nolimits_{o \neq c}  \sum\nolimits_{j} 
			\exp({ \mathbf{f}_a^c}^\top {\mathbf{f}_j^o} )
		}.
		\end{equation}
		\vspace{-0.2cm} 
		%%\vspace{-0.5cm}
	%\end{center}
	% $p(\mathbf{x}_i^c | \mathbf{x}_a^c)$ is the probability of $\mathbf{x}_i^c$ being identified similar to $\mathbf{x}_a^c$.
	
	%To maximise the matching probability $p(\mathbf{x}_i^c | \mathbf{x}_a^c)$, we equivalently minimise the cross entropy $ -\log p(\mathbf{x}_i^c | \mathbf{x}_a^c)$ between the predicted matching distribution and the ground-truth one (one-hot representation with probability one on the matching positive point), which is illustrated in Figure~\ref{fig:ICE_visualisation}.
	
	We remark: (1) Dot product measures the similarity between two vectors; (2) Eq.~(\ref{equation:ice_prob_pos}) represents the probability of a query matching a positive while Eq.~(\ref{equation:cce_prob}) is the probability of a query matching its ground-truth class.  
	%Here, the matching probability is computed by  dot product (similarity metric of two vectors) followed by softmax normalisation.
	To maximise $p(\mathbf{x}_i^c | \mathbf{x}_a^c)$ and minimise $p(\mathbf{x}_j^o | \mathbf{x}_a^c,\mathbf{x}_i^c)$ simultaneously, we minimise 
	the Kullback-Leibler divergence \citep{kullback1951information} between the predicted and ground-truth distributions,
	%
	%the KL-divergence from a predicted matching distribution to its ground-truth one, 
	%
	which is equivalent to minimise their cross entropy. 
	Since the ground-truth distribution is one-hot encoded,   
	the cross-entropy is $ -\log p(\mathbf{x}_i^c | \mathbf{x}_a^c)$. 
	%when only one positive point is considered. 
	%The distribution dimension is 1 (positive) plus the number of negatives in the mini-batch as shown in Figure~\ref{fig:ICE}. 
	
	\vspace{-0.06cm}
	To be more general, for the given anchor $\mathbf{x}_a^c$,  there may exist multiple matching points when $N_c > 2$, i.e., $|\{\mathbf{x}_i^c\}_{i \neq a}|=N_c-1>1$. 
	In this case, we predict one matching distribution per positive point.
	Our goal is to maximise the joint probability of all positive instances being correctly identified, i.e.,  
	$\label{equation:ice_prob_prod}
	p_{\mathbf{x}_a^c} = \prod\nolimits_{i \neq a} p(\mathbf{x}_i^c | \mathbf{x}_a^c)$.
	A case of two positives matching a given query is described in Figure~\ref{fig:ICE}. 
	
	%	\begin{equation}
	%	\label{equation:ice_prob_prod}
	%	p_{\mathbf{x}_a^c} = \prod\nolimits_{i \neq a} p(\mathbf{x}_i^c | \mathbf{x}_a^c). 
	%	\end{equation}
	%	
	%\begin{equation}
	%\label{equation:ice_prob_prod}
	%\begin{aligned}
	%p_{\mathbf{x}_a^c} &= \prod_{i \neq a} p({z}_i^c | \mathbf{x}_a^c) \\
	%&= \prod_{i \neq a} \frac{\exp( {\mathbf{f}_a^c}^\top {\mathbf{f}_i^c} )}
	%{	\exp( {\mathbf{f}_a^c}^\top {\mathbf{f}_i^c} ) + 
	%	\sum\limits_{o \neq c}  \sum\limits_{j} 
	%	\exp({ \mathbf{f}_a^c}^\top {\mathbf{f}_j^o} )
	%}.
	%\end{aligned}
	%\end{equation}
	
	\vspace{-0.06cm}
	In terms of mini-batch, each image in $\mathbf{X}$ serves as the anchor iteratively and we aim to maximise the joint probability of all queries $\{\{p_{\mathbf{x}_a^c}\}_{a=1}^{N_c} \}_{c=1}^C$. %Therefore, our overall objective is to obtain an error free (perfect) recognition of the mini-batch data.  
	Equivalently, we can achieve this by minimising the sum of all negative log-likelihoods. Therefore, our proposed ICE on $\mathbf{X}$ is as follows:
	%\begin{center}
		%%\vspace{-0.75cm}
		\vspace{-0.1cm}
		\begin{equation}
		\begin{aligned}
		\label{equation:ice_loss}
		L_{\mathrm{ICE}}(\mathbf{X};f) &=  -\sum\nolimits_{c=1}^C \sum\nolimits_{a=1}^{N_c} \log p_{\mathbf{x}_a^c} 
		&= -\sum\nolimits_{c=1}^C \sum\nolimits_{a=1}^{N_c} \sum\nolimits_{i \neq a} \log p(\mathbf{x}_i^c | \mathbf{x}_a^c)
		.
		\end{aligned}
		\vspace{-0.06cm}
		\end{equation}
	
	\vspace{-0.16cm}
	\subsection{Regularisation by $L_2$ Feature Normalisation}
	\label{sec:L2regularisation}
	\vspace{-0.16cm}
	Following the common practice in existing DML methods, we apply $L_2$-normalisation to feature embeddings before the inner product. 
	Therefore, the inner product denotes the cosine similarity. 
	%\citep{song2017deep,law2017deep,movshovitz2017no}.
	
	The similarity between two feature vectors is determined by their norms and the angle between them. 
	%The angle between two vectors is bounded while their norms are unbounded without $L_2$ feature normalisation. 
	Without $L_2$ normalisation, the feature norm can be very large, making the model training unstable and difficult. 
	With $L_2$ normalisation, \textit{all features are projected to a unit hypersphere surface}.
	Consequently, the semantic similarity score is merely determined by the direction of learned representations. 
	Therefore, $L_2$ normalisation can be regarded as a regulariser during training\footnote{The training without $L_2$ feature normalisation leads to the norm of features becoming very large easily and the dot product becoming INF.}.   
	Note that the principle is quite different from recent hyperspherical learning methods % based on modified CCE
	\citep{liu2017sphereface,wang2018cosface,wang2018additive,liu2017deep,liu2018decoupled,liu2018learning}. They enforce the learned \textit{weight parameters} to a unit hypersphere surface and diversify their angles. % for regularisation. 
	In contrast, feature normalisation is \textit{output regularisation} and invariant to the parametrisation of the underlying neural network \citep{pereyra2017regularizing}. 
	%In summary, by adding $L_2$ feature normalisation as a regularizer, our learning objective becomes:
	In summary, our learning objective is:
	\vspace{-0.1cm} 
	\begin{equation}
	\label{equation:ice_object_regu}
	\argmax_{f}
	\prod\nolimits_{c=1}^C \prod\nolimits_{a=1}^{N_c} p_{\mathbf{x}_a^c}
	=
	\argmin_{f} L_{\mathrm{ICE}}(\mathbf{X};f) \quad  s.t. \quad \forall a, c,  ||\mathbf{f}_a^c||_2=1.
	\end{equation}
	\vspace{-0.2cm}
	%\vskip -0.1in 
	
	The feature $L_2$-normalisation layer is implemented according to \cite{wang2017normface}. It is a differentiable layer and can be easily inserted at the output of a neural net.
	
	\vspace{-0.16cm}
	\subsection{Sample Reweighting of ICE}
	\label{sec:implict_weighting_generalisation}
	\vspace{-0.16cm}
	\textbf{Intrinsic sample weighting.}
	We find that ICE emphasises more on harder samples from the perspective of gradient magnitude.  
	%The difficulty of data points is defined by the gap to reach an error free objective. 
	We demonstrate this by deriving the partial derivatives of $L_{\mathrm{ICE}}(\mathbf{X};f)$ with respect to positive and negative examples.

	%In this section, we derive the partial derivatives of $L_{\mathrm{ICE}}(\mathbf{X};f)$ with respect to positive and negative samples. Consequently, we discover that ICE implicitly emphasizes more on harder samples. The hardness degree of data points is based on the gap to reach an error free objective. 
	
	Given the query $\mathbf{x}_a^c$, the partial derivative of its any positive instance is derived by the chain rule:
	\vspace{-0.1cm}
	\begin{equation}
	\label{equation:ice_derivative_one_pos}
	\begin{aligned}
	\frac{\partial L_{\mathrm{ICE}}(\mathbf{X};f)}{\partial \mathbf{f}_i^c} 
	%&=  - \frac{1}{p(\mathbf{x}_i^c | \mathbf{x}_a^c)} \frac{\partial p(\mathbf{x}_i^c | \mathbf{x}_a^c)}{\partial \mathbf{f}_i^c} \\
	&=- \frac{\mathbf{f}_a^c \cdot \sum\nolimits_{o \neq c}  \sum\nolimits_{j} 
		\exp({ \mathbf{f}_a^c}^\top {\mathbf{f}_j^o})  }
	{	\exp( {\mathbf{f}_a^c}^\top {\mathbf{f}_i^c} ) + 
		\sum\nolimits_{o \neq c}  \sum\nolimits_{j} 
		\exp({ \mathbf{f}_a^c}^\top {\mathbf{f}_j^o} )
	}
	&= - \mathbf{f}_a^c \cdot (1-p(\mathbf{x}_i^c | \mathbf{x}_a^c)).
	\end{aligned}
	\end{equation}
	%\vskip -0.1in 
	Since $||\mathbf{f}_a^c||_2 =1$, $w_{(\mathbf{x}_i^c;\mathbf{x}_a^c)} =||\frac{\partial L_{\mathrm{ICE}}(\mathbf{X};f)}{\partial \mathbf{f}_i^c}||_2 =(1-p(\mathbf{x}_i^c | \mathbf{x}_a^c))$ can be viewed as the weight of $\mathbf{f}_i^c$ when the anchor is $\mathbf{x}_a^c$. 
	Thus, \textit{ICE focuses more on harder positive samples}, whose $p(\mathbf{x}_i^c | \mathbf{x}_a^c)$ is lower. 

	\vspace{-0.06cm}
	Similarly, the partial derivative of its any negative sample is: 
	%\vskip -0.18in
	\begin{equation}
	\label{equation:ice_derivative_one_neg}
	\begin{aligned}
	\frac{\partial L_{\mathrm{ICE}}(\mathbf{X};f)}{\partial \mathbf{f}_j^o} 
	%&= \sum\limits_{i \neq a} - \frac{1}{p(\mathbf{x}_i^c | \mathbf{x}_a^c)} \frac{\partial p(\mathbf{x}_i^c | \mathbf{x}_a^c)}{\partial \mathbf{f}_j^o}  \\
	&= \sum\nolimits_{i \neq a} \frac{\mathbf{f}_a^c \cdot  
		\exp({ \mathbf{f}_a^c}^\top {\mathbf{f}_j^o})  }
	{	\exp( {\mathbf{f}_a^c}^\top {\mathbf{f}_i^c} ) + 
		\sum\nolimits_{o \neq c}  \sum\nolimits_{j} 
		\exp({ \mathbf{f}_a^c}^\top {\mathbf{f}_j^o} )
	}
	&= \mathbf{f}_a^c \cdot \sum\nolimits_{i \neq a} p(\mathbf{x}_j^o | \mathbf{x}_a^c,\mathbf{x}_i^c)
	,
	\end{aligned}
	\end{equation}
	where $p(\mathbf{x}_j^o | \mathbf{x}_a^c,\mathbf{x}_i^c)$ is the matching probability between $\mathbf{x}_j^o$ and $\mathbf{x}_a^c$ given that the ground-truth example is $\mathbf{x}_i^c$.
	The weight of $\mathbf{x}_j^o$ w.r.t. $\mathbf{x}_a^c$ is: %also the norm of its partial derivative
	$w_{(\mathbf{x}_j^o;\mathbf{x}_a^c)}=||\frac{\partial L_{\mathrm{ICE}}(\mathbf{X};f)}{\partial \mathbf{f}_j^o} ||_2 = \sum\nolimits_{i \neq a} p(\mathbf{x}_j^o | \mathbf{x}_a^c,\mathbf{x}_i^c)$.
	%\begin{equation}
	%\label{equation:ice_weight_one_neg}
	%\begin{aligned}
	%w_{(\mathbf{x}_j^o;\mathbf{x}_a^c)}
	%&=
	%||\frac{\partial L_{\mathrm{ICE}}(\mathbf{X};f)}{\partial \mathbf{f}_j^o} ||_2 
	%%\\
	%%&= \sum\limits_{i \neq a} \frac{  	\exp({ \mathbf{f}_a^c}^\top {\mathbf{f}_j^o})  } {	\exp( {\mathbf{f}_a^c}^\top {\mathbf{f}_i^c} ) + 	\sum\limits_{o \neq c}  \sum\limits_{j} 	\exp({ \mathbf{f}_a^c}^\top {\mathbf{f}_j^o} )}\\
	%%&
	%= \sum\limits_{i \neq a} p(\mathbf{x}_j^o | \mathbf{x}_a^c,\mathbf{x}_i^c)
	%,
	%\end{aligned}
	%\end{equation}
	Clearly, \textit{the harder negative samples own higher matching probabilities and weights}. 
	
	\vspace{-0.06cm}
	\textbf{Relative weight analysis.}
	In general,{ the relative weight \citep{tabachnick2007using} is more notable as the exact weight will be rescaled during training}, e.g., linear post-processing by multiplying the learning rate. 
	Therefore, we analyse the relative weight between two positive points of the same anchor ($i \neq k \neq a$):\begin{equation}
	\label{equation:ice_relative_weight_pos}
	\begin{aligned}
	\frac{w_{(\mathbf{x}_i^c;\mathbf{x}_a^c)}}{w_{(\mathbf{x}_k^c;\mathbf{x}_a^c)}} &= \frac{1-p(\mathbf{x}_i^c | \mathbf{x}_a^c)}{1-p(\mathbf{x}_k^c | \mathbf{x}_a^c)} 
	&=\frac{\exp( {\mathbf{f}_a^c}^\top {\mathbf{f}_k^c} ) + 
		\sum\nolimits_{o \neq c}  \sum\nolimits_{j} 
		\exp({ \mathbf{f}_a^c}^\top {\mathbf{f}_j^o} )
	}
	{\exp( {\mathbf{f}_a^c}^\top {\mathbf{f}_i^c} ) + 
		\sum\nolimits_{o \neq c}  \sum\nolimits_{j} 
		\exp({ \mathbf{f}_a^c}^\top {\mathbf{f}_j^o} )
	}.
	\end{aligned}
	\end{equation}
	Similarly, the relative weight between two negative points of the same anchor ($o \neq c,  l \neq c$) is:
	\begin{equation}
	\label{equation:ice_relative_weight_neg}
	\begin{aligned}
	\frac{w_{(\mathbf{x}_j^o;\mathbf{x}_a^c)}}
	{w_{(\mathbf{x}_k^l;\mathbf{x}_a^c)}} 
	=
	\frac{\sum\nolimits_{i \neq a} p(\mathbf{x}_j^o | \mathbf{x}_a^c,\mathbf{x}_i^c)}
	{\sum\nolimits_{i \neq a} p(\mathbf{x}_k^l | \mathbf{x}_a^c,\mathbf{x}_i^c)}
	=
	\frac{
		\exp({ \mathbf{f}_a^c}^\top {\mathbf{f}_j^o})
	}
	{
		\exp({ \mathbf{f}_a^c}^\top {\mathbf{f}_k^l})
	}.
	\end{aligned}
	\end{equation}
	Note that the positive relative weight in Eq.~(\ref{equation:ice_relative_weight_pos}) is \textit{only decided} by ${\mathbf{f}_a^c}^\top {\mathbf{f}_i^c}$ and ${\mathbf{f}_a^c}^\top {\mathbf{f}_k^c}$ while the negative relative weight in Eq.~(\ref{equation:ice_relative_weight_neg}) is \textit{only determined} by ${ \mathbf{f}_a^c}^\top {\mathbf{f}_j^o}$ and ${ \mathbf{f}_a^c}^\top {\mathbf{f}_k^l}$.
	%Furthermore, \textit{the dot product between $L_2$ normalised vectors is in the range $[-1, 1]$, which is too strictly bounded}. 
	The relative weight is merely determined by the dot product, which is in the range of $[-1, 1]$ and strictly bounded.
	
	\vspace{-0.06cm}
	\textbf{Non-linear scaling for controlling the relative weight.} 
	%When we compute compatibility, the dot product between $L_2$ normalised vectors is in the range $[-1, 1]$, which is too strictly bounded. 
	Inspired by \citep{hinton2015distilling}, we introduce a scaling parameter to modify the absolute weight non-linearly:
	%,wang2018cosface,wang2018additive,wu2018improving
	%In general, $s>=1$. Formally:
	%%\vskip -0.3in 
	%\vskip -0.15in 
	\begin{equation}
	\begin{aligned}
	\hat{w}_{(\mathbf{x}_i^c;\mathbf{x}_a^c)}
	&= \frac{ \sum\nolimits_{o \neq c}  \sum\nolimits_{j} 
		\exp(s \cdot { \mathbf{f}_a^c}^\top {\mathbf{f}_j^o})  }
	{	\exp(s \cdot  {\mathbf{f}_a^c}^\top {\mathbf{f}_i^c} ) + 
		\sum\nolimits_{o \neq c}  \sum\nolimits_{j} 
		\exp(s \cdot { \mathbf{f}_a^c}^\top {\mathbf{f}_j^o} )
	}
	&= 1-\hat{p}{(\mathbf{x}_i^c|\mathbf{x}_a^c)},
	\end{aligned}
	\end{equation}
	%\vskip -0.15in 
	\begin{equation}
	\label{equation:ice_weight_one_neg}
	\begin{aligned}
	\hat{w}_{(\mathbf{x}_j^o;\mathbf{x}_a^c)}
	&= \sum\nolimits_{i \neq a} \frac{  
		\exp(s \cdot  { \mathbf{f}_a^c}^\top {\mathbf{f}_j^o})  }
	{	\exp(s \cdot   {\mathbf{f}_a^c}^\top {\mathbf{f}_i^c} ) + 
		\sum\nolimits_{o \neq c}  \sum\nolimits_{j} 
		\exp(s \cdot  { \mathbf{f}_a^c}^\top {\mathbf{f}_j^o} )
	}
	&=\sum\nolimits_{i \neq a}
	\hat{p}(\mathbf{x}_j^o | \mathbf{x}_a^c,\mathbf{x}_i^c)
	,
	\end{aligned}
	\end{equation}
	where $s \ge 1$ is the scaling parameter.  In contrast to $p$ and $w$, $\hat{p}$ and $\hat{w}$ represent the rescaled matching probability and partial derivative weight, respectively. We remark that we scale the absolute weight non-linearly, which is an indirect way of controlling the relative weight. 
	We do not modify the relative weight directly and Eq.~(\ref{equation:ice_relative_weight_pos}) and   Eq.~(\ref{equation:ice_relative_weight_neg}) are only for introducing our motivation.  
	%%\vskip -0.16in
	
	\vspace{-0.08cm}
	Our objective is to maximise an anchor's matching probability with its any positive instance competing against its negative set.
	Therefore, we normalise the rescaled weights based on each anchor:
	%Mathematically:
	%%\vskip -0.15in 
	\begin{equation}
	\vspace{-0.16cm}
	\label{equation:final_weight_pos}
	\begin{aligned}
	\bar{w}_{(\mathbf{x}_i^c;\mathbf{x}_a^c)}
	&= \frac{1}{N} \cdot
	\frac{\hat{w}_{(\mathbf{x}_i^c;\mathbf{x}_a^c)}
	}
	{	\sum\nolimits_{i \neq a} 
		\hat{w}_{(\mathbf{x}_i^c;\mathbf{x}_a^c)}
		+ \sum\nolimits_{o \neq c}  \sum\nolimits_{j}
		\hat{w}_{(\mathbf{x}_j^o;\mathbf{x}_a^c)}
	}
	&= \frac{1}{2N} \cdot
	\frac{
		1-\hat{p}{(\mathbf{x}_i^c|\mathbf{x}_a^c)}
	}
	{	\sum\nolimits_{i \neq a} 
		(1-\hat{p}{(\mathbf{x}_i^c|\mathbf{x}_a^c)})	
	}
	,
	\end{aligned}
	%%%\vspace{-0.33cm}	
	\end{equation}
	\begin{equation}
	\label{equation:final_weight_neg}
	\begin{aligned}
	\bar{w}_{(\mathbf{x}_j^o;\mathbf{x}_a^c)}
	&= \frac{1}{N} \cdot
	\frac{\hat{w}_{(\mathbf{x}_j^o;\mathbf{x}_a^c)}}
	{	\sum\nolimits_{i \neq a} 
		\hat{w}_{(\mathbf{x}_i^c;\mathbf{x}_a^c)}
		+ \sum\nolimits_{o \neq c}  \sum\nolimits_{j}
		\hat{w}_{(\mathbf{x}_j^o;\mathbf{x}_a^c)}
	}
	&= \frac{1}{2N} \cdot
	\frac{
		\sum\nolimits_{i \neq a}
		\hat{p}(\mathbf{x}_j^o | \mathbf{x}_a^c,\mathbf{x}_i^c)
	}
	{	\sum\nolimits_{i \neq a} 
		(1-\hat{p}{(\mathbf{x}_i^c|\mathbf{x}_a^c)})	
	}
	.
	\end{aligned}
	\end{equation}
	Note that the denominators in Eq.~(\ref{equation:final_weight_pos}) and (\ref{equation:final_weight_neg}) are the accumulated weights of positives and negatives w.r.t. $\mathbf{x}_a^c$, respectively.
	\textit{Although there are much more negatives than positives, the negative set and positive set contribute equally as a whole, as indicated by $1/2$. }
	$N=\sum_{c=1}^{C} N_c$ is the total number of instances in $\mathbf{X}$.
	We select each instance as the anchor iteratively and treat all anchors equally, as indicated by $1/N$.

	%	As indicated in Eq.~(\ref{equation:final_weight_pos}) and Eq.~(\ref{equation:final_weight_neg}), the weights of positive and negative points align well with our intuition. On the one hand, for a more difficult positive instance, its rescaled probability $\hat{p}$ is less than 1 and weight is higher. On the other hand, a more difficult negative example has higher matching probability and weight.
	%	% 
	%	By introducing the scaling parameter $s$, we can \textit{control the emphasis degree on harder data points in different applications}. 

	%%%%%%%%%%%%%%%%%%%%%%%%%%%%%%%%%%%%%%%%
	\begin{algorithm}[!t]
		\caption{Learn by minimising ICE stochastically}
		\label{algorithm:ICE}
		%%%\vspace{-0.15cm}	
		\begin{algorithmic}
			\STATE \textbf{Batch setting}: $C$ classes, $N_c$ images from $c$-th class, batch size $N=\sum_{c=1}^{C} N_c$.
			\STATE \textbf{Hyper-setting}: The scaling parameter $s$ and the number of iterations $\tau$.   
			\STATE \textbf{Input}: Initialised embedding function $f$, iteration counter $iter=0$.
			%the learning rate $\gamma$.  
			\STATE \textbf{Output}: Updated $f$.
			
			\FOR{$iter < \tau$}
			
			\STATE $iter= iter + 1$. \\Sample one mini-batch randomly  $\mathbf{X} =  \{\{\mathbf{x}_i^c\}_{i=1}^{N_c}\}_{c=1}^C$.
			
			\STATE \textbf{Step 1}: Feedforward $\mathbf{X}$ into $f$ to obtain feature representations $\{\{\mathbf{f}_i^c\}_{i=1}^{N_c}\}_{c=1}^C$. 
			
			\STATE \textbf{Step 2}: Compute the similarities between an anchor and the remaining instances. Every example serves as the anchor iteratively. 
			\FOR{$\mathbf{f}_a^c \in 
				\{\{\mathbf{f}_i^c\}_{i=1}^{N_c}\}_{c=1}^C$}
			\FOR{$\mathbf{f}_i^c \in 
				\{\mathbf{f}_i^c\}_{i \neq a}$
			}  
			\STATE Compute  $p(\mathbf{x}_i^c | \mathbf{x}_a^c)$ using Eq.~(\ref{equation:ice_prob_pos}). // We do not need to compute Eq.~(\ref{equation:ice_prob_neg}).
			\ENDFOR \\
			%Compute  $p_{\mathbf{x}_a^c}$ using Eq.~(\ref{equation:ice_prob_prod}).	
			\ENDFOR
			\STATE Compute $L_{\mathrm{ICE}}(\mathbf{X};f)$ using Eq.~(\ref{equation:ice_loss}).

			\STATE \textbf{Step 3}: Gradient back-propagation to update the parameters of $f$ using  Eq.~(\ref{equation:ice_derivative_one_pos_final}). %and Eq.~(\ref{equation:ice_derivative_one_neg_final}).	
			%\STATE $\nabla_f=\partial L_\mathrm{ICE}(\mathbf{X};f)/\partial f$
			%\STATE $f=f- \gamma \cdot \nabla_f$
			\ENDFOR
		\end{algorithmic}
		%%%\vspace{-0.05cm}	
	\end{algorithm}
	%%%%%%%%%%%%%%%%%%%%%%%%%%%%%%%%%%%%%%%%%%%%%%%%%%%%
	
	\vspace{-0.08cm}
	It is worth noting that during back-propagation, the magnitudes of partial derivatives in Eq.~(\ref{equation:ice_derivative_one_pos}) and Eq.~(\ref{equation:ice_derivative_one_neg}), i.e.,
	$w_{(\mathbf{x}_i^c;\mathbf{x}_a^c)}$ and $w_{(\mathbf{x}_i^c;\mathbf{x}_a^c)}$, are replaced by 
	$\bar{w}_{(\mathbf{x}_i^c;\mathbf{x}_a^c)}$ and $\bar{w}_{(\mathbf{x}_i^c;\mathbf{x}_a^c)}$ respectively. The direction of each individual partial derivative is unchanged. However, since weights are rescaled non-linearly, the final partial derivative of each sample is changed to a better weighted combination of multiple partial derivatives. 
	{Final partial derivatives} of $L_{\mathrm{ICE}}(\mathbf{X};f)$ w.r.t. positives and negatives are: 
	%\vspace{-0.2cm} 
	\begin{equation}
	\label{equation:ice_derivative_one_pos_final}
	\begin{aligned}
	\frac{\partial L_{\mathrm{ICE}}(\mathbf{X};f)}{\partial \mathbf{f}_i^c} 
	&= - \mathbf{f}_a^c \cdot \bar{w}_{(\mathbf{x}_i^c;\mathbf{x}_a^c)}
	\text{~~ and ~~}
	\frac{\partial L_{\mathrm{ICE}}(\mathbf{X};f)}{\partial \mathbf{f}_j^o} 
	&= \mathbf{f}_a^c \cdot \bar{w}_{(\mathbf{x}_j^o;\mathbf{x}_a^c)}.
	\end{aligned}
	\end{equation}
	\vspace{-0.3cm} 
	%\vskip -0.1in 
	%	\begin{equation}
	%	\label{equation:ice_derivative_one_neg_final}
	%	\begin{aligned}
	%	\frac{\partial L_{\mathrm{ICE}}(\mathbf{X};f)}{\partial \mathbf{f}_j^o} 
	%	&= \mathbf{f}_a^c \cdot \bar{w}_{(\mathbf{x}_j^o;\mathbf{x}_a^c)}.
	%	\end{aligned}
	%	\end{equation}
	\vspace{-0.16cm}
	\subsection{A Case Study and Intuitive Explanation of ICE}
	\vspace{-0.16cm}
	To make it more clear and intuitive for understanding, we now analyse a naive case of ICE, where there are two samples per class in every mini-batch, i.e., $\forall c, Nc = 2$, $|\{\mathbf{x}_i^c\}_{i \neq a}|=N_c-1=1$. 
	In this case, for each anchor (query), there is only one positive among the remaining data points. 
	As a result, the weighting schemes in Eq.~(\ref{equation:final_weight_pos}) for positives and Eq.~(\ref{equation:final_weight_neg}) for negatives can be simplified: 
	\vspace{-0.15cm} 
	\begin{equation}
	\label{equation:final_weight_pos_naive}
	\begin{aligned}
	\bar{w}_{(\mathbf{x}_i^c;\mathbf{x}_a^c)}
	&= \frac{1}{2N} \cdot
	\frac{
		1-\hat{p}{(\mathbf{x}_i^c|\mathbf{x}_a^c)}
	}
	{	\sum\nolimits_{i \neq a} 
		(1-\hat{p}{(\mathbf{x}_i^c|\mathbf{x}_a^c)})	
	}
	= \frac{1}{N} \cdot \frac{1}{2}
	,
	\end{aligned}
	\end{equation}
	\vspace{-0.2cm} 
	\begin{equation}
	\label{equation:final_weight_neg_naive}
	\begin{aligned}
	\bar{w}_{(\mathbf{x}_j^o;\mathbf{x}_a^c)}
	&= \frac{1}{2N} \cdot
	\frac{
		\sum\nolimits_{i \neq a}
		\hat{p}(\mathbf{x}_j^o | \mathbf{x}_a^c,\mathbf{x}_i^c)
	}
	{	\sum\nolimits_{i \neq a} 
		(1-\hat{p}{(\mathbf{x}_i^c|\mathbf{x}_a^c)})	
	}
	&= \frac{1}{N} \cdot \frac{1}{2}\cdot
	\frac{
		\hat{p}(\mathbf{x}_j^o | \mathbf{x}_a^c,\mathbf{x}_i^c)
	}
	{ 1-\hat{p}{(\mathbf{x}_i^c|\mathbf{x}_a^c)}	
	}.
	\end{aligned}
	\end{equation}
	%\textbf{Interpretation.}  
	%It is intuitive to understand the weighing generalisation scheme from Eq.~(\ref{equation:final_weight_pos_naive}) and Eq.~(\ref{equation:final_weight_neg_naive}).
	Firstly, we have $N$ anchors that are treated equally as indicated by $1/N$. Secondly, for each anchor, we aim to recognise its positive example correctly. However, there is a \textit{sample imbalance problem} because \textit{each anchor has only one positive and many negatives}. ICE addresses it by treating the positive set (single point) and negative set (multiple points) equally, i.e., $1/2$ in Eq.~(\ref{equation:final_weight_pos_naive}) and Eq.~(\ref{equation:final_weight_neg_naive})
	\footnote{The weight sum of negatives: $\sum_{o \neq c} \sum_{j} \hat{p}(\mathbf{x}_j^o | \mathbf{x}_a^c,\mathbf{x}_i^c) = 1-\hat{p}{(\mathbf{x}_i^c|\mathbf{x}_a^c)} =>
		\sum_{o \neq c} \sum_{j} \bar{w}_{(\mathbf{x}_j^o;\mathbf{x}_a^c)} = \bar{w}_{(\mathbf{x}_i^c;\mathbf{x}_a^c)} = 1/(2N)
		$.
	}. 
	Finally, as there are many negative samples, we aim to focus more on informative ones, i.e., harder negative instances with higher matching probabilities with a given anchor. 
	%This is achieved by our rescaled probability $\hat{p}$, which is a non-linear transformation of the original probability $p$. 
	The non-linear transformation can help control the relative weight between two negative points. 
	
	The weighting scheme shares the same principle as the popular temperature-based categorical cross entropy \citep{hinton2015distilling,oreshkin2018tadam}. 
	The key is that we should consider not only focusing on harder examples, but also the emphasis degree.
	%, which is adjusted by the scaling parameter. 

	\vspace{-0.16cm}
	\subsection{Complexity Analysis}
	\label{sec:complexity_analysis}
	\vspace{-0.16cm} 
	Algorithm~\ref{algorithm:ICE} summarises the learning process with ICE.
	As presented there, the input data format of ICE is the same as CCE, i.e., images and their corresponding labels. 
	In contrast to other methods which require rigid input formats \citep{schroff2015facenet,sohn2016improved}, e.g., triplets and n-pair tuplets, 
	ICE is much more flexible. 
	We iteratively select one image as the anchor. For each anchor, we aim to maximise its matching probabilities with its positive samples against its negative examples. Therefore, \textit{the computational complexity over one mini-batch is} $O(N^2)$, being the same as recent online metric learning approaches \citep{song2016deep,wang2019deep}. Note that in FaceNet \citep{schroff2015facenet} and $N$-pair-mc \citep{sohn2016improved}, expensive sample mining and class mining are applied, respectively. 
	
	\vspace{-0.1cm}
	\section{Experiments}
	\vspace{-0.1cm}
	\begin{table}
		\parbox{.55\textwidth}{
			\captionsetup{width=.55\textwidth}
			\caption{
				A summary of three fine-grained datasets. % in the context of cars, birds, and products respectively. 
				Training and test classes are disjoint. `\#' refers to the number of each item.   
				There are only 5.3 images per class on average in SOP. 
			}
			\label{table:datasets}
			\vspace{-0.35cm} 
			\begin{center}
				\begin{small}
					%\begin{sc}
					\setlength{\tabcolsep}{2.8pt} % Default value: 6pt
					%%\vspace{-0.1cm}
					\begin{tabular}{lccc}
						\toprule
						Datasets & CARS196 & CUB-200-2011  & SOP  \\
						\midrule
						Context & Cars & Birds & Products \\
						\#Total classes & 196 & 200 & 22,634 \\
						\#Total images & 16,185 & 11,788 & 120,053 \\
						\#Training classes & 98 & 100 & 11,318 \\
						\#Training images & 8,054 & 5,864 &  59,551\\
						\#Test classes & 98 & 100 & 11,316 \\
						\#Test images & 8,131 & 5,924 & 60,502 \\
						%\#Average images per class & 82.6 & 58.9 & 5.3\\
						\bottomrule
					\end{tabular}
					%\end{sc}
				\end{small}
			\end{center}
		}
		\hfill
		\parbox{.41\textwidth}{
			%%\vspace{-0.46cm}
			\caption{
				The results of different reweighting parameters $s$ on SOP in terms of Recall@$K$ (\%).  
				There are 90 classes and 2 images per class in a mini-batch, i.e., the batch size is 180.
				%$180=90 \times 2$ indicates the batch size $N=180$ and there are 90 classes and 2 images per class in a mini-batch. 
			}
			\label{table:s}
			\vspace{-0.42cm}
			\begin{center}
				\begin{small}
					%\begin{sc}
					\setlength{\tabcolsep}{6pt} % Default value: 6pt
					%%\vspace{-0.1cm}
					\begin{tabular}{lccc}
						\toprule
						Reweighting & R@1 & R@10 & R@100 \\
						\midrule
						$s=1$ & 42.0 & 58.1 & 74.1 \\
						$s=16$ & 71.0 & 85.6 & 93.8 \\
						$s=32$ & 73.6 & 87.5 & 94.7 \\
						$s=48$ & {76.9} & 89.7 & {95.5} \\
						$s=64$ & \textbf{77.3} & \textbf{90.0} & \textbf{95.6} \\
						$s=80$ & 75.4 & 88.7 & 94.9 \\
						\bottomrule
					\end{tabular}
					%\end{sc}
				\end{small}
			\end{center}
		}
		\vspace{-0.3cm}
	\end{table}

	%
	%%%%%%%%%%%%%%%%%%%%%%%%%%%%%%%%
	% Note use of \abovespace and \belowspace to get reasonable spacing
	% above and below tabular lines.
	%	\begin{table}[!t]
	%		\caption{
	%			A summary of three fine-grained datasets. % in the context of cars, birds, and products respectively. 
	%			Training and test classes are disjoint on each dataset. `\#' refers to the number of each item.   
	%			There are only 5.3 images per class on average in SOP. 
	%		}
	%		%\vskip -0.06in
	%		\label{table:datasets}
	%		\begin{center}
	%			\begin{small}
	%				%\begin{sc}
	%					\setlength{\tabcolsep}{1.2pt} % Default value: 6pt
	%					
	%					\begin{tabular}{lccc}
	%						\toprule
	%						Datasets & CARS196 & CUB-200-2011  & SOP  \\
	%						\midrule
	%						Context & Cars & Birds & Products \\
	%						\#Total classes & 196 & 200 & 22,634 \\
	%						\#Total images & 16,185 & 11,788 & 120,053 \\
	%						\#Training classes & 98 & 100 & 11,318 \\
	%						\#Training images & 8,054 & 5,864 &  59,551\\
	%						\#Test classes & 98 & 100 & 11,316 \\
	%						\#Test images & 8,131 & 5,924 & 60,502 \\
	%						\#Average images per class & 82.6 & 58.9 & 5.3
	%						\\
	%						\bottomrule
	%					\end{tabular}
	%				%\end{sc}
	%			\end{small}
	%		\end{center}
	%		%\vskip -0.2in
	%	\end{table}
	%%%%%%%%%%%%%%%%%%%%%%%%%%%%%%%%%\\
	\vspace{-0.1cm}
	\subsection{Implementation Details and Evaluation Settings}
	\vspace{-0.1cm}
	For data augmentation and preprocessing, we follow \citep{song2016deep,song2017deep}.  
	In detail, we first resize the input images to $256 \times 256$ and then crop it at $227 \times 227$. We use random cropping and horizontal mirroring for data augmentation during training. To fairly compare with the results reported in \citep{song2017deep}, we use a centre cropping without horizontal flipping in the test phase. 
	For the embedding size, we set it to 512 on all datasets following \citep{sohn2016improved,law2017deep,wang2019ranked}.
	To compare fairly with \citep{song2017deep,law2017deep,movshovitz2017no}, we choose GoogLeNet V2 (with batch normalisation) \citep{ioffe2015batch} as the backbone architecture initialised by the publicly available pretrained model on ImageNet \citep{russakovsky2015imagenet}. We simply change the original 1000-neuron fully connected layers followed by softmax normalisation and CCE to 512-neuron fully connected layers followed by the proposed ICE. For faster convergence, we randomly initialise the new layers and optimise them with 10 times larger learning rate than the others as in \citep{song2016deep}.
	
	\vspace{-0.06cm}
	We implement our algorithm in the Caffe framework \citep{jia2014caffe}. The source code will be available soon. 
	
	\vspace{-0.06cm}
	\textbf{Datasets.}
	Following the evaluation protocol in \citep{song2016deep,song2017deep}, we test our proposed method on three popular fine-grained datasets including CARS196 \citep{krause20133d}, CUB-200-2011 \citep{wah2011caltech} and SOP \citep{song2016deep}. A summary of the datasets is given in Table~\ref{table:datasets}. We also keep the same train/test splits. We remark that to test the generalisation and transfer capability of the learned deep metric, the training and test classes are disjoint.

	\vspace{-0.06cm}
	\textbf{Evaluation protocol.}
	%We evaluate the learned representations on two different tasks, i.e., the image retrieval performance in terms of Recall@$K$ \citep{song2016deep} and the image clustering quality in terms of Normalised Mutual Information (NMI) \citep{schutze2008introduction}.
	We evaluate the learned representations on the image retrieval task in terms of Recall@$K$ performance \citep{song2016deep}.
	Given a query, its $K$ nearest neighbours are retrieved from the database. Its retrieval score is one if there is an image of the same class in the $K$ nearest neighbours and zero otherwise. Recall@$K$ is the average score of all queries. 
	
	\vspace{-0.06cm}
	\textbf{Training settings.} All the experiments are run on a single PC equipped with Tesla V100 GPU with 32GB RAM. For optimisation, we use the stochastic gradient descent (SGD) with a weight decay of $1e^{-5}$ and a momentum of 0.8. The base learning rate is set as $1e^{-3}$. %on SOP.
	The training converges at $20k$ iterations on SOP while 
	$4k$ iterations on CARS196 and CUB-200-2011.
	As for the hyper-parameters, we study their impacts in Sec. \ref{sec:ablation_study} and supplementary material.
	The mini-batch size is 60 for small datasets CARS196 and CUB-200-2011 while 180 for the large benchmark SOP. 
	Additionally, we set $C=6,  N_c=10$ on CARS196 and CUB-200-2011 while $C=90, N_c=2$ on SOP. 
	The design reasons are: 1) SOP has only 5.3 images per class on average. Therefore $N_c$ cannot be very large; 2) It helps to simulate the global structure of deep embeddings, where the database is large and only a few matching instances exist. 
	
	\vspace{-0.06cm}
	The analysis of batch content, batch size and embedding size is presented in the supplementary material.

	%%%%%%%%%%%%%%%%%%%%%%%%%%%%%%%%
	% Note use of \abovespace and \belowspace to get reasonable spacing
	% above and below tabular lines.
	\begin{table*}[!t]
		%%%\vspace{-0.38cm}
		\caption{Comparison with the state-of-the-art methods on CARS196, CUB-200-2011 and SOP in terms of Recall@$K$ (\%).
			% and NMI (\%). 
			All the compared methods use GoogLeNet V2 as the backbone architecture. `--' means the results which are not reported in the original paper. 
			The best results in the first block using single embedding are bolded.   
		}
		\vspace{-0.2cm}
		\label{table:SOTA}
		\begin{center}
			\begin{small}
				%\begin{sc}
				\setlength{\tabcolsep}{6.0pt} % Default value: 6pt
				%%\vspace{-0.35cm}
				\begin{tabular}{lcccc|cccc|ccc}
					\toprule
					& \multicolumn{4}{c|}{CARS196} & \multicolumn{4}{c|}{CUB-200-2011}
					& \multicolumn{3}{c}{SOP} \\ 
					\cmidrule(r){2-12}
					$K$ & 1 & 2 & 4 & 8  & 1 & 2 & 4 & 8  & 1 & 10 & 100  \\
					\midrule
					Without fine-tuning & 35.6 & 47.3 & 59.4 & 72.2 & 40.1 & 53.2 & 66.0 & 76.6 & 43.7 & 60.8 & 76.5 \\
					Fine-tuned with CCE & 48.8 & 58.5 & 71.0 & 78.4 &  46.0 & 58.0 & 69.3 & 78.3 &  51.7 & 69.8 & 85.3  \\
					%
					%\midrule
					%Contrastive \citep{bell2015learning} & 21.7 & 32.3 & 46.1 & 58.9 & 72.2 & 83.4 & 26.4 & 37.7 & 49.8 & 62.3 & 76.4 & 85.3 \\
					Triplet Semihard  & 51.5 & 63.8 & 73.5 & 82.4   & 42.6 & 55.0 & 66.4 & 77.2  & 66.7 & 82.4 & 91.9 \\
					Lifted Struct & 53.0 & 65.7 & 76.0 & 84.3  & 43.6 & 56.6 & 68.6 & 79.6   & 62.5 & 80.8 & 91.9\\
					%Binomial Deviance \citep{ustinova2016learning} & -- & -- & -- & -- & -- & -- & 52.8 & 64.4 & 74. 7 & 83.9 & 90.4 & 94. 3 \\
					%Histogram \citep{ustinova2016learning} & -- & -- & -- & -- & -- & -- & 50.3 & 61.9 & 72.6 & 82.4 & 88.8 & 93.7 \\
					
					N-pair-mc  & 53.9 & 66.8 & 77.8 & 86.4  & 45.4 & 58.4 & 69.5 & 79.5  & 66.4 & 83.2 & 93.0 \\

					%HDC* \citep{yuan2017hard} & 73.7 & 83.2 & 89.5 & 93.8 & 96.7 & 98.4 & 53.6 & 65.7 & 77.0 & 85.6 & 91.5 & 95.5 \\
					%BIER* \citep{opitz2017bier} & 78.0 & 85.8 & 91.1 & 95.1 & 97.3 & 98.7 & 55.3 & 67.2 & 76.9 & 85.1 & 91.7 & 95.5 \\

					%Margin \citep{wu2017sampling} & 79.6 & 86.5 & 91.9 & 95.1 & 97.3 & -- & \textbf{\textit{63.6}} & 74.4 & 83.1 & 90.0 & 94.2 & -- \\

					Struct Clust  & 58.1 & 70.6 & 80.3 & 87.8 & 48.2 & 61.4 & 71.8 & 81.9 & 67.0 & 83.7 & 93.2   \\
					
					Spectral Clust & 73.1 & 82.2 & 89.0 & 93.0 & 53.2 & 66.1 & 76.7 & 85.3 & 67.6 & 83.7 & 93.3 \\

					%Proxy Triplet \citep{movshovitz2017no} & 55.9 & 68.0 & 74.0 & 78.0 & 54.4  &  -- & -- & -- & -- & --   \\
					
					Proxy NCA & {73.2} & 82.4 & 86.4 & 88.7 &  49.2 & 61.9 & 67.9 & 72.4  & 73.7 & -- & -- \\
					
					%HTL \citep{ge2018deep} & \textbf{81.4} & \textbf{88.0} & \textbf{92.7} & \textbf{95.7} & -- & \textbf{57.1} & \textbf{68.8} & \textbf{78.7} & \textbf{86.5} & -- & \textbf{74.8} & \textbf{88.3} & \textbf{94.8} & -- \\ 
					
					%\midrule
					RLL & 74.0 & 83.6 & 90.1& 94.1& 57.4 &\textbf{69.7}& {79.2}& \textbf{86.9} & 76.1& 89.1& 95.4  \\
					ICE & \textbf{77.0} & \textbf{85.3} & \textbf{91.3} & \textbf{94.8}   &
					
					\textbf{58.3} & {69.5} & \textbf{{79.4}} & {86.7}   & 
					
					\textbf{77.3} & \textbf{90.0} & \textbf{95.6} \\
					\midrule
					RLL-(L,M,H)  & 82.1 & 89.3 & {93.7} & {96.7} &  61.3 & {72.7} & {82.7} & {89.4}  &  79.8 & 91.3 & 96.3 \\
					ICE-(L, M, H) & {82.8} & {89.5} & {93.7} & {96.4}  & {61.4} & {73.2} & {82.5} & {89.2}  & {80.1} & {91.8} & {96.6} \\
					
					%\textbf{RLL-(L,M,H)} & \textbf{81.8} & \textbf{89.1} & \textbf{93.9} & \textbf{96.6} & \textbf{--}  &
					%\textbf{61.3} & \textbf{72.7} & \textbf{82.7} & \textbf{89.4} & \textbf{66.1} & 
					%\textbf{79.8} & \textbf{91.3} & \textbf{96.3} & 90.4\\
					
					%PDDM+Triplet \citep{huang2016local} & 46.4 & 58.2 & 70.3 & 80.1 & 88.6 & 92.6 & 50.9 & 62.1 & 73.2 & 82.5 & 91.1 & 94.4 \\
					%PDDM+Quadruplet \citep{huang2016local} & 57.4 & 68.6 & 80.1 & 89.4 & 92.3 & 94.9 & 58.3 & 69.2 & 79.0 & 88.4 & 93.1 & 95.7 \\
					%HDC* \citep{yuan2017hard} & 83.8 & 89.8 & 93.6 & 96.2 & 97.8 & 98.9 & 60.7 & 72.4 & 81.9 & 89.2 & 93.7 & 96.8 \\   
					%BIER* \citep{opitz2017bier} &\textbf{ 87.2} & 92.2 & 95.3 & 97.4 & 98.5 & 99.3 & 63.7 & 74.0 & 82.5 & 89.3 & 93.8 & 96.8 \\
					%Margin \citep{wu2017sampling} & 86.9 & 92.7 & 95.6 & 97.6 & 98.7 & -- & \textbf{63.9} & 75.3 & 84.4 & 90.6 & 94.8 & -- \\
					\bottomrule
				\end{tabular}
				%\end{sc}
			\end{small}
		\end{center}
		%\vspace{-0.3cm}
	\end{table*}
	%%%%%%%%%%%%%%%%%%%%%%%%%%%%%%%%%
	\vspace{-0.16cm}
	\subsection{Quantitative Results}
	\vspace{-0.16cm}
	\textbf{Remarks.} For a fair comparison, we remark that the methods group \citep{ustinova2016learning, harwood2017smart,wang2017deep,duan2018deep,lin2018deep,suh2019stochastic,zheng2019hardness} using GoogLeNet V1 \citep{szegedy2015going} and another group \citep{wu2017sampling,cakir2019deep, sanakoyeu2019divide} using ResNet-50 \citep{he2016deep} are not benchmarked. Besides, ensemble models \citep{yuan2017hard,opitz2017bier,kim2018attention,xuan2018deep} are not considered.
	% since ours is a single model. 
	%
	%We do not compare with HTL~\citep{ge2018deep} because it constructs a hierarchical similarity tree \textit{among the whole training set} and updates the tree after every epoch, which is highly unscalable and expensive in terms of both computation and memory.
	HTL~\citep{ge2018deep} also uses GoogLeNet V2, but it constructs a hierarchical similarity tree {over the whole training set} and updates the tree every epoch, thus being highly unscalable and expensive in terms of both computation and memory.
	That is why HTL achieves better performance on small datasets but performs worse than ours on the large dataset SOP.
	Finally, there are some other orthogonal deep metric learning research topics that are worth studying together in the future, e.g., a robust distance metric \citep{yuan2019signal} and metric learning with continuous labels \citep{kim2019deep}.
	In GoogLeNet V2, there are three fully connected layers of different depth.  
	We refer them based on their depth: L for
	the low-level layer (inception-3c/output), M for the mid-level layer (inception-4e/output) and H for the high-level layer (inception5b/output).  
	By default, we use only `H'. We also report the results of their combination (L, M, H) for reference following RLL \citep{wang2019ranked}.  
	%	\footnote{
	%		
	%		% -- not representative of true deep metric learning.}. 
	%		%It achieves good performance on small datasets but is not comparable with ours on the large dataset SOP.
	%	}
	
	\textbf{Competitors.} All the compared baselines, Triplet Semihard \citep{schroff2015facenet}, Lifted Struct \citep{song2016deep}, $N$-pair-mc \citep{sohn2016improved}, Struct Clust \citep{song2017deep}, Spectral Clust \citep{law2017deep}, Proxy-NCA \citep{movshovitz2017no}, RLL \citep{wang2019ranked} and our ICE are trained and evaluated using the same settings: 
	(1) GoogLeNet V2 serves as the backbone network; 
	(2) All models are initialised with the same pretrained model on ImageNet; (3) All apply the same data augmentation during training and use a centre-cropped image during testing. The results of some baselines \citep{schroff2015facenet,song2016deep,sohn2016improved} are from \citep{song2017deep}, which means they are reimplemented there for a fair comparison. 
	In addition, the results of vanilla GoogLeNet V2 pretrained on ImageNet without fine-tuning and with fine-tuning via minimising CCE are reported in \citep{law2017deep}, which can be regarded as the most basic baselines.
	%because our proposed ICE is inspired by CCE.
	%
	Among these baselines, Proxy NCA is not scalable as class-level proxies are learned during training. 
	Struct Clust and Spectral Clust are clustering-motivated methods which explicitly aim to optimise the clustering quality. 
	%
	%In addition to the unscalable Proxy NCA, both of them work very well in the small datasets CARS196 and CUB-200-2011 but insignificant in the large dataset SOP. More specifically, Spectral Clust is close to the state-of-the-art in CARS 196 and CUB-200-2011 but performs much worse in SOP than Proxy-NCA. The main reason is that the SOP test set has a large number of classes (11,316) but only  5.3 images per class in average. As a result, the K-means clustering algorithm \citep{kanungo2002efficient} is hard to generalise. Therefore, NMI metric is less reliable for SOP as mentioned in \citep{law2017deep}. 
	We highlight that clustering performance Normalised Mutual Information (NMI) \citep{schutze2008introduction} is not a good assessment for SOP \citep{law2017deep} because SOP has a large number of classes but only  5.3 images per class on average. Therefore, we only report and compare Recall@$K$ performance. 
	
	\vspace{-0.06cm}
	\textbf{Results.} Table~\ref{table:SOTA} compares the results of our ICE and those of the state-of-the-art DML losses. ICE achieves the best Recall@1 performance on all benchmarks. We observe that  only RLL achieves comparable performance in a few terms. However, RLL is more complex since it has three hyper-parameters in total: one weight scaling parameter and two distance margins for positives and negatives, respectively.
	In addition, its perspective is different since it processes the positive set together similarly with \citep{triantafillou2017few,wang2019multi}.   
	We note that \citep{wang2019multi} is also complex in designing weighting schemes and contains four control hyper-parameters. However, our Recall@$1$ on SOP is 77.3\%, which is only 0.9\% lower than 78.2\% of \citep{wang2019multi}.   
	%Our ICE achieves state-of-the-art Recall@$1$ performance on all datasets. We also obtain the best NMI results on CARS196 and CUB-200-2011 except SOP, where the K-means is hard to generalise \citep{law2017deep}.
	%In more detail, Spectral Clust is the current state-of-the-art on CUB-200-2011. We outperform it by 4.1\% and 3.1\% in terms of Recall@$1$ and NMI, respectively. 
	%In terms of Recall@$1$, our performance on SOP is higher than that of the state-of-the-art Proxy NCA by about 2\%.    
	%
	It is also worth mentioning that among these approaches, except fine-tuned models with CCE, only our method has a clear probability interpretation and aims to maximise the joint instance-level matching probability. 
	As observed, apart from being unscalable, CCE's performance is much worse than the state-of-the-art methods. Therefore, ICE can be regarded as a successful exploration of softmax regression for learning deep representations in DML. 
	%
	%\subsection{Qualitative Results}
	%\textbf{Qualitative results.}
	The t-SNE visualisation \citep{van2014accelerating} of learned embeddings are available in the supplementary material.  
	\vspace{-0.16cm}
	\subsection{Analysis of Sample Reweighting in ICE}
	\label{sec:ablation_study}
	\vspace{-0.16cm}
	%\subsubsection{Weight scaling parameter}
	%%%\vspace{-0.15cm}
	We empirically study the impact of the weight scaling parameter $s$, which is the only hyper-parameter of ICE. It functions similarly with the popular sample mining or example weighting \citep{wang2019ranked,wang2019deep,wang2019multi} widely applied in the baselines in Table~\ref{table:SOTA}. 
	Generally, different $s$ corresponds to different emphasis degree on difficult examples. 
	When $s$ is larger, more difficult instances are assigned with relatively higher weights.
	%In more detail, when $s=0$, all hard (informative) and easy (non-informative) samples are treated equally. When $s$ is larger, more difficult instances are assigned with relatively higher weights. 
	%which is an essential property of sample mining and many loss functions. 
	
	\vspace{-0.06cm}
	In general, small datasets are more sensitive to minor changes of hyper-settings and much easier to overfit. Therefore, the experiments are conducted on the large dataset SOP. The results are shown in Table~\ref{table:s}. Note that when $s$ is too small, e.g., $s=1$, we observe that the training does not converge, which demonstrates the necessity of weighting/mining samples.
	The most significant observation is that focusing on difficult samples is better but the emphasis degree should be properly controlled. When $s$ increases from 16 to 64, the performance grows gradually. However, when $s = 80$,  we observe the performance drops a lot. That may be because extremely hard samples, e.g., outliers, are emphasised when $s$ is too large.

	%%%%%%%%%%%%%%%%%%%%%%%%%%%%%%%%
	% Note use of \abovespace and \belowspace to get reasonable spacing
	% above and below tabular lines.
	%	\begin{table}[!t]
	%		\caption{
	%			The results of different scaling parameters $s$ on SOP in terms of Recall@$K$ (\%).  $180=90 \times 2$ indicates the batch size $N=180$ and there are 90 classes and 2 images per class in a mini-batch. 
	%		}
	%		%%\vskip -0.2in
	%		\label{table:s}
	%		\begin{center}
	%			\begin{small}
	%				%\begin{sc}
	%					\setlength{\tabcolsep}{6pt} % Default value: 6pt
	%					
	%					\begin{tabular}{lccc}
	%						\toprule
	%						$180=90 \times 2$ & R@1 & R@10 & R@100 \\
	%						\midrule
	%						$s=1$ & 23.2 & 41.1 & 65.7 \\
	%						$s=16$ & 71.0 & 85.6 & 93.8 \\
	%						$s=32$ & 73.6 & 87.5 & 94.7 \\
	%						$s=48$ & \textbf{75.8} & 88.9 & \textbf{95.2} \\
	%						$s=64$ & \textbf{75.8} & \textbf{89.0} & \textbf{95.2} \\
	%						$s=80$ & 75.4 & 88.7 & 94.9 \\
	%						\bottomrule
	%					\end{tabular}
	%				%\end{sc}
	%			\end{small}
	%		\end{center}
	%		%%\vskip -0.2in
	%	\end{table}
	%%%%%%%%%%%%%%%%%%%%%%%%%%%%%%%%%\\

	\vspace{-0.16cm}
	\section{Conclusion}
	\vspace{-0.16cm}
	In this paper, we propose a novel instance-level softmax regression framework, named instance cross entropy, for deep metric learning.
	%, which is inspired by the success of CCE. 
	%
	Firstly, the proposed ICE has clear probability interpretation and exploits structured semantic similarity information among multiple instances. 
	Secondly, ICE is scalable to infinitely many classes, which is required by DML. 
	Thirdly, ICE has only one weight scaling hyper-parameter, which works as mining informative examples and can be easily selected via cross-validation.  
	Finally, distance thresholds are not applied to achieve intraclass compactness and interclass separability. This indicates that ICE makes no assumptions about intraclass variances and the boundaries between different classes. Therefore ICE owns general applicability.  
	
	%Finally, we demonstrate its superiority on three real-world datasets.   
	
	%
	%Many improvements of CCE have been made recently \citep{liu2016large,liu2017deep,liu2018decoupled,liu2018learning,wang2018additive,wang2018cosface}. We believe that our ICE can also benefit from them. 
	%In the future, many potential improvements could be explored as many variants of CCE have been proposed recently \citep{liu2016large,liu2017deep,liu2018decoupled,liu2018learning,wang2018additive,wang2018cosface}.

	%\section{Related Work}
	
	%\subsection{Noise-aware Modelling}
	%Exploiting clean labels\\
	%Exploiting auxiliary models
	%
	%\subsection{Sample reweighting}
	%Iterative noisy labels detection and cleaning
	
	%\subsection{Robust loss functions}
	
	%\subsection{Explicit regularisations}
	%
	%Confidence penalty \citep{pereyra2017regularizing}
	%
	%Weight decay, norm
	%
	%Proper fitting \citep{ma2018dimensionality}
	
	%\subsection{Robustness against adversarial examples?}
	%Pending!
	%
	%A novel perspective for enhancing robustness against adversarial attack without any intention during training. 
	%
	%Learn from mix-up paper \& Review recent papers. 
	
	%%%%%%%%%%%%%%%%%%References
	%\todo{try to reduce the number of references}
	%\bibliographystyle{splncs}
	%\bibliography{references_nips2018}

	\bibliographystyle{iclr2020_conference}
	\bibliography{DML}
	
	\appendix
	\newpage
	\begin{center}
		\Large \textbf{Supplementary Material for \\Instance Cross Entropy for Deep Metric Learning}
	\end{center}
	
	\section{More Ablation Studies}
	%%\vspace{-0.15cm}
	\subsection{Batch Content}
	%%\vspace{-0.15cm}
	%In DML, training and testing classes are disjoint for testing the generalisation of the learned deep metric.  
	%In this case, theoretically we can make the training classes and images as many as possible. 
	%Therefore, the training set can be regarded as a collected subset (subtask). Furthermore, each mini-batch can be viewed as an even smaller subtask. 
	
	We evaluate the impact of batch content which consists of $C$ classes and $k$ images per class, i.e., $\forall c, N_c =k$. The batch size $N=C \times k$ is set to 180. In our experiments, we change the number of classes $C$ from 36 to 90, and the number of images $k$ from 2 to 5, while keeping the batch size unchanged. Table~\ref{table:batch_content} shows the results on SOP dataset. We observe that when there are more classes in the mini-batch, the performance is better. We conjecture that as the number of classes increases, the mini-batch training becomes more difficult and helps the model to generalise better.  
	
	%%%%%%%%%%%%%%%%%%%%%%%%%%%%%%%%
	% Note use of \abovespace and \belowspace to get reasonable spacing
	% above and below tabular lines.
	\begin{table}[!h]
		%\vskip -0.1in
		\caption{
			The impact of batch content $C \times k$ on SOP in terms of Recall@$K$ (\%). The batch size is $N=180$ and the scaling parameter is $s = 64$.  
			%Note that when $T=96$, we cannot train as the loss never drops, which means it is not proper to focus on extremely hard samples. 
		}
		\label{table:batch_content}
		\begin{center}
			\begin{small}
				%\begin{sc}
				\setlength{\tabcolsep}{6pt} % Default value: 6pt
				
				\begin{tabular}{lccc}
					\toprule
					$N=180, s=64$ & R@1 & R@10 & R@100 \\
					\midrule
					$C \times k=90 \times 2$ & 77.3 & 90.0 & 95.6 \\
					$C \times k=60 \times 3$ & 75.2 & 88.7 & 95.2 \\
					$C \times k=45 \times 4$ & 74.9 & 88.7 & 95.3 \\
					$C \times k=36 \times 5$ & 74.6 & 88.7 & 95.4 \\
					\bottomrule
				\end{tabular}
				%\end{sc}
			\end{small}
		\end{center}
		%\vskip -0.2in
	\end{table}
	%%%%%%%%%%%%%%%%%%%%%%%%%%%%%%%%%\\
	
	%We keep the batch size To do this, we set the batch size
	%We fix the batch size as $N=180$ and evaluate the impact of batch content (the number of classes -- $C$ and the number of images per class -- $k$). We fix the image number per class as , i.e., $\forall c, N_c =k$.
	%We conduct the experiments on the large dataset SOP and show the results in Table~\ref{table:batch_content}. 
	%We observe that when there are more classes in the mini-batch, the performance is better. We conjecture that as the number of classes increases, the subtask becomes more difficult and helps the model to generalise better. 
	
	%It is intuitive that the batch size $N$ and batch content  are hyper-parameters that are independent of specific DML algorithms. 
	%
	%In the literature, the batch size and content are set empirically and differently in different approaches \citep{song2017deep,law2017deep}. 
	%Here, we conduct experiments on the large dataset SOP to evaluate their impacts on the recognition performance. 

	%%\vspace{-0.15cm}
	\subsection{Batch Size}
	%%\vspace{-0.15cm}
	
	To explore different batch size $N$, we fix $k=2$ and only change $C$. In this case, $N=C\times 2$. Table~\ref{table:batch_size} shows that as the number of classes increases, the performance grows. In detail, when the number of classes increases from 50 to 90, the performance raises from 74.4\% to 77.3\% accordingly. One reason may be that as the number of classes increases, it fits the global structure of the test set better, where there are a large number of classes but only a few positive examples. In addition, the increasing difficulty of mini-batch training can help the model to generalise better.

	%%%%%%%%%%%%%%%%%%%%%%%%%%%%%%%%
	% Note use of \abovespace and \belowspace to get reasonable spacing
	% above and below tabular lines.
	\begin{table}[!h]
		%\vskip -0.1in
		\caption{
			The results of different batch size $N$ on SOP in terms of Recall@$K$ (\%). While changing $C$, we fix $k=2$ and $s = 64$. Therefore, $N=C\times2$.  
			%Note that when $T=96$, we cannot train as the loss never drops, which means it is not proper to focus on extremely hard samples. 
		}
		%%\vskip -0.2in
		\label{table:batch_size}
		\begin{center}
			\begin{small}
				%\begin{sc}
				\setlength{\tabcolsep}{6pt} % Default value: 6pt
				
				\begin{tabular}{lccc}
					\toprule
					$k=2, s=64$ & R@1 & R@10 & R@100 \\
					\midrule
					%$N=190$ & 75.9 & 89.0 & 95.1 \\
					$N=180$ & 77.3 & 90.0 & 95.6 \\
					%$N=170$ & 75.8 & 89.0 & 95.1 \\
					$N=160$ & 75.4 & 88.8 & 95.1 \\
					%$N=150$ & 75.6 & 89.0 & 95.1 \\
					$N=140$ & 75.1 & 88.7 & 95.2 \\
					$N=120$ & 75.1 & 88.6 & 95.2 \\
					$N=100$ & 74.4 & 88.2 & 95.1 \\
					\bottomrule
				\end{tabular}
				%\end{sc}
			\end{small}
		\end{center}
		%\vskip -0.2in
	\end{table}
	%%%%%%%%%%%%%%%%%%%%%%%%%%%%%%%%%\\
	
	%%\vspace{-0.15cm}
	\subsection{Embedding Size}
	%%\vspace{-0.15cm}
	The dimension of feature representations is an important factor in many DML methods. We conduct experiments on SOP to see the influence of different embedding size. The results are presented in Table \ref{table:embedding_size}. We observe that when the embedding size is very small, e.g., 64, the performance is much worse. 
	The performance increases gradually as the embedding size grows.
	% enough e.g., from 512 to 1024,  is neglectable. 

	%%%%%%%%%%%%%%%%%%%%%%%%%%%%%%%%
	% Note use of \abovespace and \belowspace to get reasonable spacing
	% above and below tabular lines.
	\begin{table}[!h]
		%\vskip -0.1in
		\caption{
			The results of different embedding size on SOP in terms of Recall@$K$ (\%). In all experiments: $s = 64$, $C=90, k =2$. $N=C\times k =90\times2$.  
			%Note that when $T=96$, we cannot train as the loss never drops, which means it is not proper to focus on extremely hard samples. 
		}
		%%\vskip -0.2in
		\label{table:embedding_size}
		\begin{center}
			\begin{small}
				%\begin{sc}
				\setlength{\tabcolsep}{6pt} % Default value: 6pt
				
				\begin{tabular}{lccc}
					\toprule
					$180=90\times 2, s=64$ & R@1 & R@10 & R@100 \\
					\midrule
					$64$ & 72.6 & 87.1 & 94.0 \\
					$128$ & 74.3 & 87.9 & 94.5 \\
					$256$ & 75.2 & 88.6 & 94.8 \\
					$512$ &  77.3 & 90.0 & 95.6 \\
					%$1024$ & 76.0 & 89.1 & 95.3 \\
					\bottomrule
				\end{tabular}
				%\end{sc}
			\end{small}
		\end{center}
		%\vskip -0.2in
	\end{table}

	\section{t-SNE Visualisation}
\begin{figure}[!h]
		%\vskip 0.2in
		\centering
		\centerline{\includegraphics[width=\columnwidth]{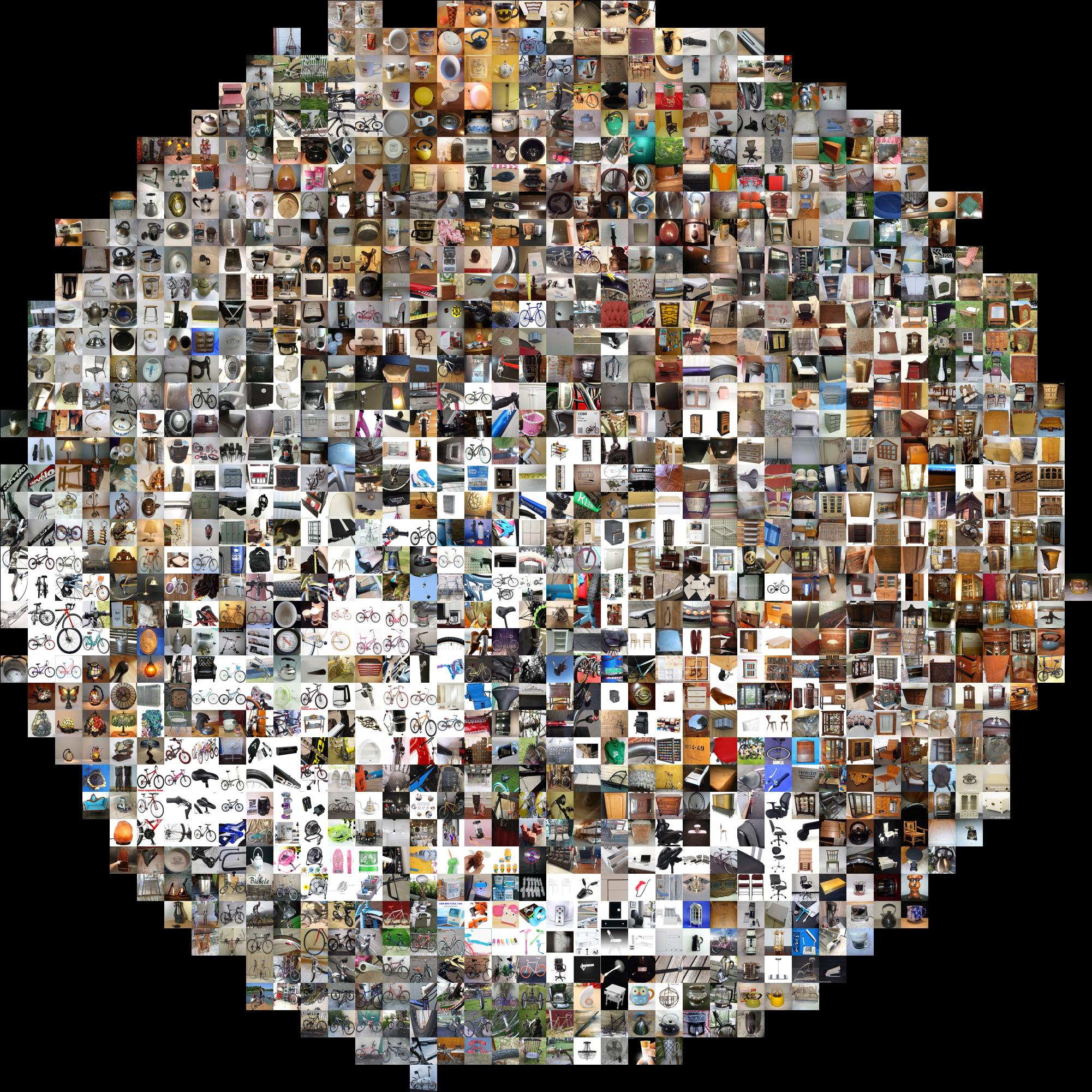}}
		\caption{t-SNE visualisation \citep{van2014accelerating} on the SOP test set. Best viewed on a monitor when zoomed in.}
		%\vskip -0.2in
		\label{fig:SOP_plot}
	\end{figure}

\begin{figure}[!h]
		%\vskip 0.2in
		\centering
		\centerline{\includegraphics[width=\columnwidth]{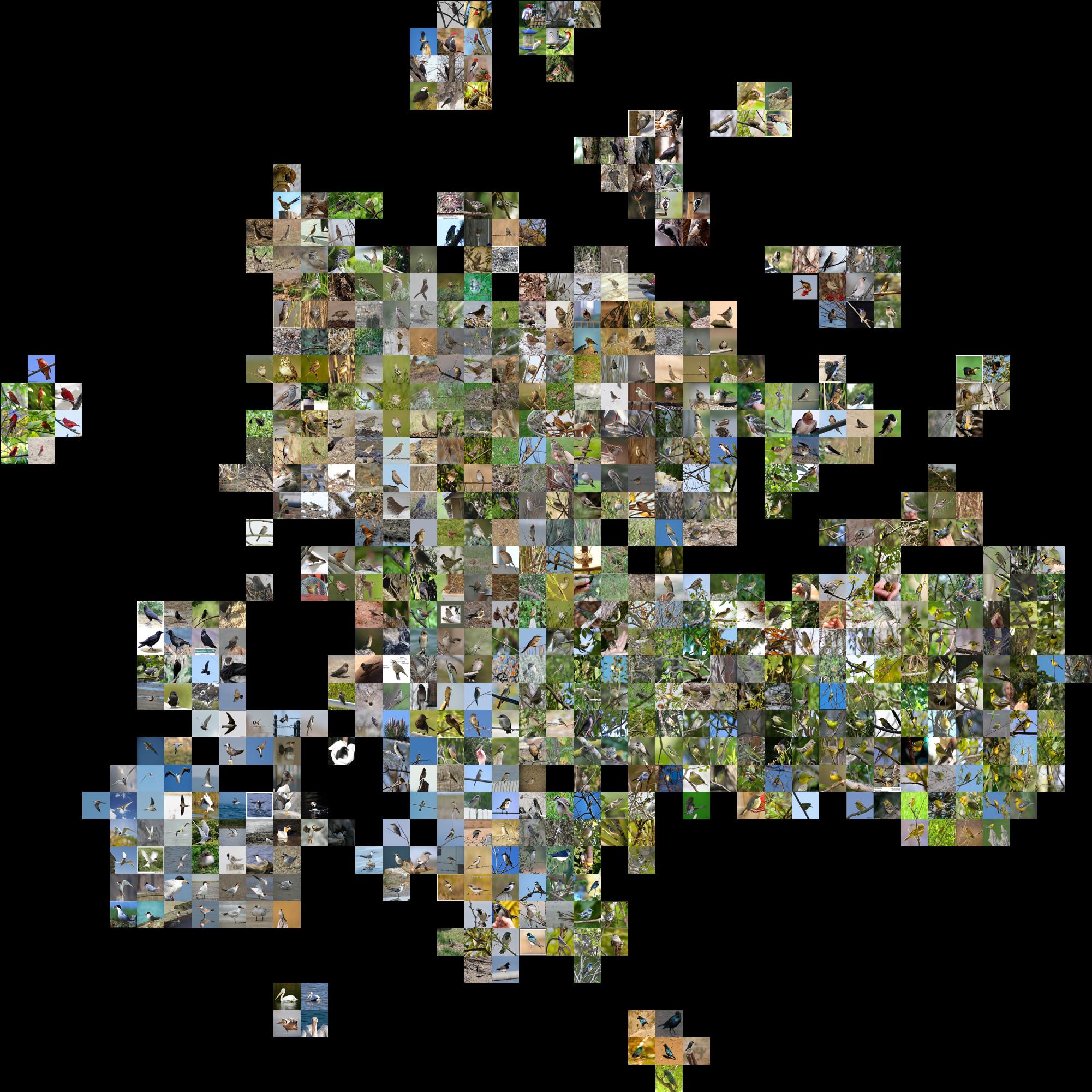}}
		\caption{t-SNE visualisation \citep{van2014accelerating} on the CUB-200-2011 test set. Best viewed on a monitor when zoomed in.}
		%\vskip -0.2in
		\label{fig:CUB_plot}
	\end{figure}

\begin{figure}[!h]
		%\vskip 0.2in
		\centering
		\centerline{\includegraphics[width=\columnwidth]{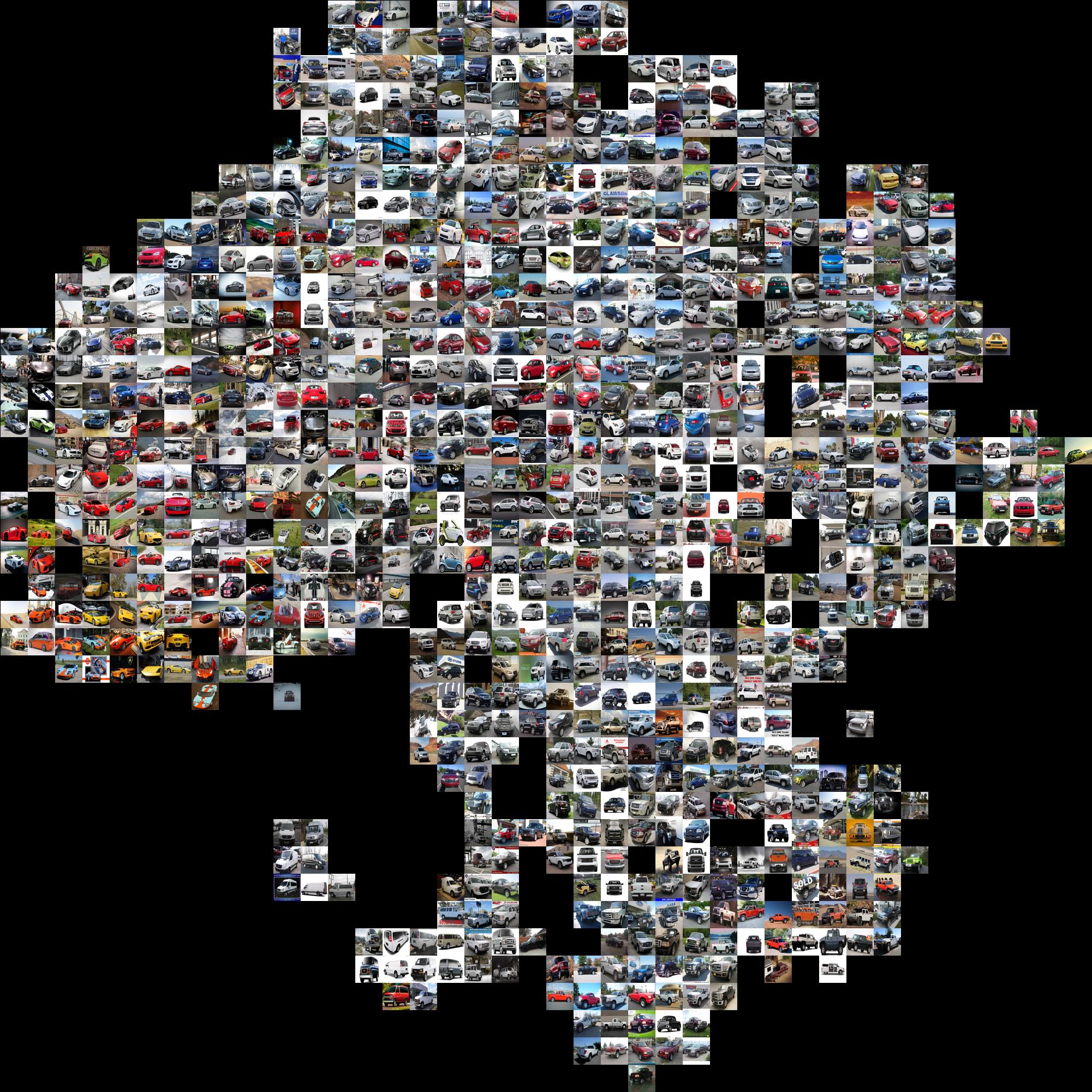}}
		\caption{t-SNE visualisation \citep{van2014accelerating} on the CARS196 test set. Best viewed on a monitor when zoomed in.}
		%\vskip -0.2in
		\label{fig:CARS_plot}
	\end{figure}

	The t-SNE visualisation \citep{van2014accelerating} of learned embeddings is available in Figures \ref{fig:SOP_plot}, \ref{fig:CUB_plot}, \ref{fig:CARS_plot}.  
	
	
\end{document}

%% file: math_commands.tex
\usepackage{amsmath,amsfonts,bm}

\def\eqref#1{equation~\ref{#1}}
\def\1{\bm{1}}

\DeclareMathAlphabet{\mathsfit}{\encodingdefault}{\sfdefault}{m}{sl}
\SetMathAlphabet{\mathsfit}{bold}{\encodingdefault}{\sfdefault}{bx}{n}

\DeclareMathOperator*{\argmax}{arg\,max}
\DeclareMathOperator*{\argmin}{arg\,min}

%% file: ICE_revision_V02.bbl
\begin{thebibliography}{62}
\providecommand{\natexlab}[1]{#1}
\providecommand{\url}[1]{\texttt{#1}}
\expandafter\ifx\csname urlstyle\endcsname\relax
  \providecommand{\doi}[1]{doi: #1}\else
  \providecommand{\doi}{doi: \begingroup \urlstyle{rm}\Url}\fi

\bibitem[Bucher et~al.(2016)Bucher, Herbin, and Jurie]{bucher2016improving}
Maxime Bucher, St{\'e}phane Herbin, and Fr{\'e}d{\'e}ric Jurie.
\newblock Improving semantic embedding consistency by metric learning for
  zero-shot classification.
\newblock In \emph{ECCV}, 2016.

\bibitem[Cakir et~al.(2019)Cakir, He, Xia, Kulis, and Sclaroff]{cakir2019deep}
Fatih Cakir, Kun He, Xide Xia, Brian Kulis, and Stan Sclaroff.
\newblock Deep metric learning to rank.
\newblock In \emph{CVPR}, 2019.

\bibitem[Chopra et~al.(2005)Chopra, Hadsell, and LeCun]{chopra2005learning}
Sumit Chopra, Raia Hadsell, and Yann LeCun.
\newblock Learning a similarity metric discriminatively, with application to
  face verification.
\newblock In \emph{CVPR}, 2005.

\bibitem[Duan et~al.(2018)Duan, Zheng, Lin, Lu, and Zhou]{duan2018deep}
Yueqi Duan, Wenzhao Zheng, Xudong Lin, Jiwen Lu, and Jie Zhou.
\newblock Deep adversarial metric learning.
\newblock In \emph{CVPR}, 2018.

\bibitem[Ge et~al.(2018)Ge, Huang, Dong, and Scott]{ge2018deep}
Weifeng Ge, Weilin Huang, Dengke Dong, and Matthew~R Scott.
\newblock Deep metric learning with hierarchical triplet loss.
\newblock In \emph{ECCV}, 2018.

\bibitem[Goldberger et~al.(2005)Goldberger, Hinton, Roweis, and
  Salakhutdinov]{goldberger2005neighbourhood}
Jacob Goldberger, Geoffrey~E Hinton, Sam~T Roweis, and Ruslan~R Salakhutdinov.
\newblock Neighbourhood components analysis.
\newblock In \emph{NeurIPS}, 2005.

\bibitem[Harwood et~al.(2017)Harwood, Kumar, Carneiro, Reid, Drummond,
  et~al.]{harwood2017smart}
Ben Harwood, BG~Kumar, Gustavo Carneiro, Ian Reid, Tom Drummond, et~al.
\newblock Smart mining for deep metric learning.
\newblock In \emph{ICCV}, 2017.

\bibitem[He et~al.(2016)He, Zhang, Ren, and Sun]{he2016deep}
Kaiming He, Xiangyu Zhang, Shaoqing Ren, and Jian Sun.
\newblock Deep residual learning for image recognition.
\newblock In \emph{CVPR}, 2016.

\bibitem[Hinton et~al.(2015)Hinton, Vinyals, and Dean]{hinton2015distilling}
Geoffrey Hinton, Oriol Vinyals, and Jeffrey Dean.
\newblock Distilling the knowledge in a neural network.
\newblock In \emph{NeurIPS Deep Learning and Representation Learning Workshop},
  2015.

\bibitem[Huang et~al.(2016)Huang, Loy, and Tang]{huang2016local}
Chen Huang, Chen~Change Loy, and Xiaoou Tang.
\newblock Local similarity-aware deep feature embedding.
\newblock In \emph{NeurIPS}, 2016.

\bibitem[Ioffe \& Szegedy(2015)Ioffe and Szegedy]{ioffe2015batch}
Sergey Ioffe and Christian Szegedy.
\newblock Batch normalization: Accelerating deep network training by reducing
  internal covariate shift.
\newblock In \emph{ICML}, 2015.

\bibitem[Jia et~al.(2014)Jia, Shelhamer, Donahue, Karayev, Long, Girshick,
  Guadarrama, and Darrell]{jia2014caffe}
Yangqing Jia, Evan Shelhamer, Jeff Donahue, Sergey Karayev, Jonathan Long, Ross
  Girshick, Sergio Guadarrama, and Trevor Darrell.
\newblock Caffe: Convolutional architecture for fast feature embedding.
\newblock In \emph{ACMMM}, 2014.

\bibitem[Kim et~al.(2019)Kim, Seo, Laptev, Cho, and Kwak]{kim2019deep}
Sungyeon Kim, Minkyo Seo, Ivan Laptev, Minsu Cho, and Suha Kwak.
\newblock Deep metric learning beyond binary supervision.
\newblock In \emph{CVPR}, 2019.

\bibitem[Kim et~al.(2018)Kim, Goyal, Chawla, Lee, and Kwon]{kim2018attention}
Wonsik Kim, Bhavya Goyal, Kunal Chawla, Jungmin Lee, and Keunjoo Kwon.
\newblock Attention-based ensemble for deep metric learning.
\newblock In \emph{ECCV}, 2018.

\bibitem[Krause et~al.(2013)Krause, Stark, Deng, and Fei-Fei]{krause20133d}
Jonathan Krause, Michael Stark, Jia Deng, and Li~Fei-Fei.
\newblock 3d object representations for fine-grained categorization.
\newblock In \emph{ICCV Workshop}, 2013.

\bibitem[Kullback \& Leibler(1951)Kullback and
  Leibler]{kullback1951information}
Solomon Kullback and Richard~A Leibler.
\newblock On information and sufficiency.
\newblock \emph{The Annals of Mathematical Statistics}, pp.\  79--86, 1951.

\bibitem[Law et~al.(2017)Law, Urtasun, and Zemel]{law2017deep}
Marc~T Law, Raquel Urtasun, and Richard~S Zemel.
\newblock Deep spectral clustering learning.
\newblock In \emph{ICML}, 2017.

\bibitem[Law et~al.(2019)Law, Snell, Farahmand, Urtasun, and
  Zemel]{law2019dimensionality}
Marc~T Law, Jake Snell, Amir-massoud Farahmand, Raquel Urtasun, and Richard~S
  Zemel.
\newblock Dimensionality reduction for representing the knowledge of
  probabilistic models.
\newblock In \emph{ICLR}, 2019.

\bibitem[Lin et~al.(2018)Lin, Duan, Dong, Lu, and Zhou]{lin2018deep}
Xudong Lin, Yueqi Duan, Qiyuan Dong, Jiwen Lu, and Jie Zhou.
\newblock Deep variational metric learning.
\newblock In \emph{ECCV}, 2018.

\bibitem[Liu et~al.(2016)Liu, Wen, Yu, and Yang]{liu2016large}
Weiyang Liu, Yandong Wen, Zhiding Yu, and Meng Yang.
\newblock Large-margin softmax loss for convolutional neural networks.
\newblock In \emph{ICML}, 2016.

\bibitem[Liu et~al.(2017{\natexlab{a}})Liu, Wen, Yu, Li, Raj, and
  Song]{liu2017sphereface}
Weiyang Liu, Yandong Wen, Zhiding Yu, Ming Li, Bhiksha Raj, and Le~Song.
\newblock Sphereface: Deep hypersphere embedding for face recognition.
\newblock In \emph{CVPR}, 2017{\natexlab{a}}.

\bibitem[Liu et~al.(2017{\natexlab{b}})Liu, Zhang, Li, Yu, Dai, Zhao, and
  Song]{liu2017deep}
Weiyang Liu, Yan-Ming Zhang, Xingguo Li, Zhiding Yu, Bo~Dai, Tuo Zhao, and
  Le~Song.
\newblock Deep hyperspherical learning.
\newblock In \emph{NeurIPS}, 2017{\natexlab{b}}.

\bibitem[Liu et~al.(2018{\natexlab{a}})Liu, Lin, Liu, Liu, Yu, Dai, and
  Song]{liu2018learning}
Weiyang Liu, Rongmei Lin, Zhen Liu, Lixin Liu, Zhiding Yu, Bo~Dai, and Le~Song.
\newblock Learning towards minimum hyperspherical energy.
\newblock In \emph{NeurIPS}, 2018{\natexlab{a}}.

\bibitem[Liu et~al.(2018{\natexlab{b}})Liu, Liu, Yu, Dai, Lin, Wang, Rehg, and
  Song]{liu2018decoupled}
Weiyang Liu, Zhen Liu, Zhiding Yu, Bo~Dai, Rongmei Lin, Yisen Wang, James~M
  Rehg, and Le~Song.
\newblock Decoupled networks.
\newblock In \emph{CVPR}, 2018{\natexlab{b}}.

\bibitem[Movshovitz-Attias et~al.(2017)Movshovitz-Attias, Toshev, Leung, Ioffe,
  and Singh]{movshovitz2017no}
Yair Movshovitz-Attias, Alexander Toshev, Thomas~K Leung, Sergey Ioffe, and
  Saurabh Singh.
\newblock No fuss distance metric learning using proxies.
\newblock In \emph{ICCV}, 2017.

\bibitem[Nickel \& Kiela(2018)Nickel and Kiela]{nickel2018learning}
Maximillian Nickel and Douwe Kiela.
\newblock Learning continuous hierarchies in the lorentz model of hyperbolic
  geometry.
\newblock In \emph{ICML}, 2018.

\bibitem[Opitz et~al.(2017)Opitz, Waltner, Possegger, and
  Bischof]{opitz2017bier}
Michael Opitz, Georg Waltner, Horst Possegger, and Horst Bischof.
\newblock Bier-boosting independent embeddings robustly.
\newblock In \emph{ICCV}, 2017.

\bibitem[Oreshkin et~al.(2018)Oreshkin, L{\'o}pez, and
  Lacoste]{oreshkin2018tadam}
Boris Oreshkin, Pau~Rodr{\'\i}guez L{\'o}pez, and Alexandre Lacoste.
\newblock Tadam: Task dependent adaptive metric for improved few-shot learning.
\newblock In \emph{NeurIPS}, 2018.

\bibitem[Pereyra et~al.(2017)Pereyra, Tucker, Chorowski, Kaiser, and
  Hinton]{pereyra2017regularizing}
Gabriel Pereyra, George Tucker, Jan Chorowski, {\L}ukasz Kaiser, and Geoffrey
  Hinton.
\newblock Regularizing neural networks by penalizing confident output
  distributions.
\newblock In \emph{ICLR Workshop}, 2017.

\bibitem[Russakovsky et~al.(2015)Russakovsky, Deng, Su, Krause, Satheesh, Ma,
  Huang, Karpathy, Khosla, Bernstein, et~al.]{russakovsky2015imagenet}
Olga Russakovsky, Jia Deng, Hao Su, Jonathan Krause, Sanjeev Satheesh, Sean Ma,
  Zhiheng Huang, Andrej Karpathy, Aditya Khosla, Michael Bernstein, et~al.
\newblock Imagenet large scale visual recognition challenge.
\newblock \emph{International Journal of Computer Vision}, pp.\  211--252,
  2015.

\bibitem[Sanakoyeu et~al.(2019)Sanakoyeu, Tschernezki, Buchler, and
  Ommer]{sanakoyeu2019divide}
Artsiom Sanakoyeu, Vadim Tschernezki, Uta Buchler, and Bjorn Ommer.
\newblock Divide and conquer the embedding space for metric learning.
\newblock In \emph{CVPR}, 2019.

\bibitem[Schroff et~al.(2015)Schroff, Kalenichenko, and
  Philbin]{schroff2015facenet}
Florian Schroff, Dmitry Kalenichenko, and James Philbin.
\newblock Facenet: A unified embedding for face recognition and clustering.
\newblock In \emph{CVPR}, 2015.

\bibitem[Sch{\"u}tze et~al.(2008)Sch{\"u}tze, Manning, and
  Raghavan]{schutze2008introduction}
Hinrich Sch{\"u}tze, Christopher~D Manning, and Prabhakar Raghavan.
\newblock \emph{Introduction to information retrieval}.
\newblock Cambridge University Press, 2008.

\bibitem[Shi et~al.(2016)Shi, Yang, Zhu, Liao, Lei, Zheng, and
  Li]{shi2016embedding}
Hailin Shi, Yang Yang, Xiangyu Zhu, Shengcai Liao, Zhen Lei, Weishi Zheng, and
  Stan~Z Li.
\newblock Embedding deep metric for person re-identification: A study against
  large variations.
\newblock In \emph{ECCV}, 2016.

\bibitem[Snell et~al.(2017)Snell, Swersky, and Zemel]{snell2017prototypical}
Jake Snell, Kevin Swersky, and Richard Zemel.
\newblock Prototypical networks for few-shot learning.
\newblock In \emph{NeurIPS}, 2017.

\bibitem[Sohn(2016)]{sohn2016improved}
Kihyuk Sohn.
\newblock Improved deep metric learning with multi-class n-pair loss objective.
\newblock In \emph{NeurIPS}, 2016.

\bibitem[Song et~al.(2016)Song, Xiang, Jegelka, and Savarese]{song2016deep}
Hyun~Oh Song, Yu~Xiang, Stefanie Jegelka, and Silvio Savarese.
\newblock Deep metric learning via lifted structured feature embedding.
\newblock In \emph{CVPR}, 2016.

\bibitem[Song et~al.(2017)Song, Jegelka, Rathod, and Murphy]{song2017deep}
Hyun~Oh Song, Stefanie Jegelka, Vivek Rathod, and Kevin Murphy.
\newblock Deep metric learning via facility location.
\newblock In \emph{CVPR}, 2017.

\bibitem[Suh et~al.(2019)Suh, Han, Kim, and Lee]{suh2019stochastic}
Yumin Suh, Bohyung Han, Wonsik Kim, and Kyoung~Mu Lee.
\newblock Stochastic class-based hard example mining for deep metric learning.
\newblock In \emph{CVPR}, 2019.

\bibitem[Szegedy et~al.(2015)Szegedy, Liu, Jia, Sermanet, Reed, Anguelov,
  Erhan, Vanhoucke, and Rabinovich]{szegedy2015going}
Christian Szegedy, Wei Liu, Yangqing Jia, Pierre Sermanet, Scott Reed, Dragomir
  Anguelov, Dumitru Erhan, Vincent Vanhoucke, and Andrew Rabinovich.
\newblock Going deeper with convolutions.
\newblock In \emph{CVPR}, 2015.

\bibitem[Tabachnick et~al.(2007)Tabachnick, Fidell, and
  Ullman]{tabachnick2007using}
Barbara~G Tabachnick, Linda~S Fidell, and Jodie~B Ullman.
\newblock \emph{Using multivariate statistics}.
\newblock Pearson Boston, MA, 2007.

\bibitem[Tadmor et~al.(2016)Tadmor, Rosenwein, Shalev-Shwartz, Wexler, and
  Shashua]{tadmor2016learning}
Oren Tadmor, Tal Rosenwein, Shai Shalev-Shwartz, Yonatan Wexler, and Amnon
  Shashua.
\newblock Learning a metric embedding for face recognition using the multibatch
  method.
\newblock In \emph{NeurIPS}, 2016.

\bibitem[Triantafillou et~al.(2017)Triantafillou, Zemel, and
  Urtasun]{triantafillou2017few}
Eleni Triantafillou, Richard Zemel, and Raquel Urtasun.
\newblock Few-shot learning through an information retrieval lens.
\newblock In \emph{NeurIPS}, 2017.

\bibitem[Ustinova \& Lempitsky(2016)Ustinova and
  Lempitsky]{ustinova2016learning}
Evgeniya Ustinova and Victor Lempitsky.
\newblock Learning deep embeddings with histogram loss.
\newblock In \emph{NeurIPS}, 2016.

\bibitem[Van Der~Maaten(2014)]{van2014accelerating}
Laurens Van Der~Maaten.
\newblock Accelerating t-sne using tree-based algorithms.
\newblock \emph{The Journal of Machine Learning Research}, pp.\  3221--3245,
  2014.

\bibitem[Vinyals et~al.(2016)Vinyals, Blundell, Lillicrap, Wierstra,
  et~al.]{vinyals2016matching}
Oriol Vinyals, Charles Blundell, Timothy Lillicrap, Daan Wierstra, et~al.
\newblock Matching networks for one shot learning.
\newblock In \emph{NeurIPS}, 2016.

\bibitem[Wah et~al.(2011)Wah, Branson, Welinder, Perona, and
  Belongie]{wah2011caltech}
Catherine Wah, Steve Branson, Peter Welinder, Pietro Perona, and Serge
  Belongie.
\newblock The caltech-ucsd birds-200-2011 dataset.
\newblock 2011.

\bibitem[Wang et~al.(2017{\natexlab{a}})Wang, Xiang, Cheng, and
  Yuille]{wang2017normface}
Feng Wang, Xiang Xiang, Jian Cheng, and Alan~Loddon Yuille.
\newblock Normface: l 2 hypersphere embedding for face verification.
\newblock In \emph{ACMMM}, 2017{\natexlab{a}}.

\bibitem[Wang et~al.(2018{\natexlab{a}})Wang, Cheng, Liu, and
  Liu]{wang2018additive}
Feng Wang, Jian Cheng, Weiyang Liu, and Haijun Liu.
\newblock Additive margin softmax for face verification.
\newblock In \emph{ICLR Workshop}, 2018{\natexlab{a}}.

\bibitem[Wang et~al.(2018{\natexlab{b}})Wang, Wang, Zhou, Ji, Li, Gong, Zhou,
  and Liu]{wang2018cosface}
Hao Wang, Yitong Wang, Zheng Zhou, Xing Ji, Zhifeng Li, Dihong Gong, Jingchao
  Zhou, and Wei Liu.
\newblock Cosface: Large margin cosine loss for deep face recognition.
\newblock In \emph{CVPR}, 2018{\natexlab{b}}.

\bibitem[Wang et~al.(2017{\natexlab{b}})Wang, Zhou, Wen, Liu, and
  Lin]{wang2017deep}
Jian Wang, Feng Zhou, Shilei Wen, Xiao Liu, and Yuanqing Lin.
\newblock Deep metric learning with angular loss.
\newblock In \emph{ICCV}, 2017{\natexlab{b}}.

\bibitem[Wang et~al.(2019{\natexlab{a}})Wang, Hua, Kodirov, Hu, Garnier, and
  Robertson]{wang2019ranked}
Xinshao Wang, Yang Hua, Elyor Kodirov, Guosheng Hu, Romain Garnier, and Neil~M
  Robertson.
\newblock Ranked list loss for deep metric learning.
\newblock In \emph{CVPR}, 2019{\natexlab{a}}.

\bibitem[Wang et~al.(2019{\natexlab{b}})Wang, Hua, Kodirov, Hu, and
  Robertson]{wang2019deep}
Xinshao Wang, Yang Hua, Elyor Kodirov, Guosheng Hu, and Neil~M. Robertson.
\newblock Deep metric learning by online soft mining and class-aware attention.
\newblock In \emph{AAAI}, 2019{\natexlab{b}}.

\bibitem[Wang et~al.(2019{\natexlab{c}})Wang, Han, Huang, Dong, and
  Scott]{wang2019multi}
Xun Wang, Xintong Han, Weilin Huang, Dengke Dong, and Matthew~R Scott.
\newblock Multi-similarity loss with general pair weighting for deep metric
  learning.
\newblock In \emph{CVPR}, 2019{\natexlab{c}}.

\bibitem[Wen et~al.(2016)Wen, Zhang, Li, and Qiao]{wen2016discriminative}
Yandong Wen, Kaipeng Zhang, Zhifeng Li, and Yu~Qiao.
\newblock A discriminative feature learning approach for deep face recognition.
\newblock In \emph{ECCV}, 2016.

\bibitem[Wu et~al.(2017)Wu, Manmatha, Smola, and
  Kr{\"a}henb{\"u}hl]{wu2017sampling}
Chao-Yuan Wu, R~Manmatha, Alexander~J Smola, and Philipp Kr{\"a}henb{\"u}hl.
\newblock Sampling matters in deep embedding learning.
\newblock In \emph{ICCV}, 2017.

\bibitem[Wu et~al.(2018)Wu, Efros, and Yu]{wu2018improving}
Zhirong Wu, Alexei~A Efros, and Stella~X Yu.
\newblock Improving generalization via scalable neighborhood component
  analysis.
\newblock In \emph{ECCV}, 2018.

\bibitem[Xuan et~al.(2018)Xuan, Souvenir, and Pless]{xuan2018deep}
Hong Xuan, Richard Souvenir, and Robert Pless.
\newblock Deep randomized ensembles for metric learning.
\newblock In \emph{ECCV}, 2018.

\bibitem[Yuan et~al.(2019)Yuan, Deng, Tang, Tang, and Chen]{yuan2019signal}
Tongtong Yuan, Weihong Deng, Jian Tang, Yinan Tang, and Binghui Chen.
\newblock Signal-to-noise ratio: A robust distance metric for deep metric
  learning.
\newblock In \emph{CVPR}, 2019.

\bibitem[Yuan et~al.(2017)Yuan, Yang, and Zhang]{yuan2017hard}
Yuhui Yuan, Kuiyuan Yang, and Chao Zhang.
\newblock Hard-aware deeply cascaded embedding.
\newblock In \emph{ICCV}, 2017.

\bibitem[Zhang et~al.(2018)Zhang, Yu, Karaman, Zhang, and
  Chang]{zhang2018heated}
Xu~Zhang, Felix~Xinnan Yu, Svebor Karaman, Wei Zhang, and Shih-Fu Chang.
\newblock Heated-up softmax embedding.
\newblock \emph{arXiv preprint arXiv:1809.04157}, 2018.

\bibitem[Zheng et~al.(2019)Zheng, Chen, Lu, and Zhou]{zheng2019hardness}
Wenzhao Zheng, Zhaodong Chen, Jiwen Lu, and Jie Zhou.
\newblock Hardness-aware deep metric learning.
\newblock In \emph{CVPR}, 2019.

\end{thebibliography}
